\documentclass[runningheads]{llncs}
\usepackage{graphicx}

\usepackage{algorithm}
\usepackage{algpseudocode}

\usepackage{adjustbox}
\usepackage{makecell}
\usepackage{tikz,pgf} 

\usepackage{comment}
\usepackage{amsmath,amssymb} 
\usepackage{color}
\usepackage{hyperref}
\usepackage{wrapfig}
\usepackage{booktabs}
\usepackage{graphicx}

\usetikzlibrary{positioning}
\usetikzlibrary{calc,through,backgrounds}

\usepackage[accsupp]{axessibility}  

\usepackage{booktabs}

\begin{document}
\pagestyle{headings}
\mainmatter
\def\ECCVSubNumber{3518}  

\title{FrequencyLowCut Pooling - Plug \& Play against Catastrophic Overfitting}

\titlerunning{FLC Pooling}
%
\author{Julia Grabinski\inst{1,2,3}\orcidID{0000-0002-8371-1734} \and
Steffen Jung\inst{4}\orcidID{0000-0001-8021-791X} \and
Janis Keuper\inst{2,3}\orcidID{0000-0002-1327-1243} \and \\
Margret Keuper\inst{1,4}\orcidID{0000-0002-8437-7993}}
\authorrunning{Grabinski et al.}
%
\institute{Visual Computing, Siegen University, Germany \and
Competence Center High Performance Computing, Fraunhofer ITWM, Kaiserslautern, Germany \and
Institute for Machine Learning and Analytics, Offenburg University, Germany
\and
Max Planck Institute for Informatics, Saarland Informatics Campus, Germany}

\maketitle

\begin{abstract}
Over the last years, Convolutional Neural Networks (CNNs) have been the dominating neural architecture in a wide range of computer vision tasks. From an image and signal processing point of view, this success might be a bit surprising as the inherent spatial pyramid design of most CNNs is apparently violating basic signal processing laws, i.e.~{\it Sampling Theorem} in their down-sampling operations. However, since poor sampling appeared not to affect model accuracy, this issue has been broadly neglected until model robustness started to receive more attention. Recent work \cite{grabinski2022aliasing} in the context of adversarial attacks and distribution shifts, showed after all, that there is a strong correlation between the vulnerability of CNNs and aliasing artifacts induced by poor down-sampling operations. This paper builds on these findings and introduces an aliasing free down-sampling operation which can easily be plugged into any CNN architecture: FrequencyLowCut pooling. Our experiments show, that in combination with simple and Fast Gradient Sign Method (FGSM) adversarial training, our hyper-parameter free operator substantially improves model robustness and avoids catastrophic overfitting.
Our code is available at \url{https://github.com/GeJulia/flc_pooling}
\\

\end{abstract}

\section{Introduction}
The robustness of convolutional neural networks has evolved to being one of the most crucial computer vision research topics in recent years. While state-of-the-art models provide high accuracy in many tasks, their susceptibility to adversarial attacks \cite{auto_attack} and even common corruptions \cite{hendrycks2019using} is hampering their deployment in many practical applications. Therefore, a wide range of publications aim to provide models with increased robustness by adversarial training (AT) schemes \cite{harnesssing,mart,trades}, sophisticated data augmentation techniques \cite{rebuffi2021fixing} and enriching the training with additional data \cite{carmon2019unlabeled,gowal2021uncovering}. As a result, robuster models can be learned with common CNN architectures, yet arguably at a high training cost - even without investigating the reasons for CNN's vulnerability. These reasons are of course multifold, starting with the high dimensionality of the feature space and sparse training data such that models easily tend to overfit \cite{rice2020overfitting,wong2020fast}. Recently, the pooling operation in CNNs has been discussed in a similar context for example in \cite{grabinski2022aliasing} who measured the correlation between aliasing and a network's susceptibility to adversarial attacks. \cite{zhang2019making} have shown that commonly used pooling operations even prevent the smoothness of image representations under small input translations.

Our contributions are summarized as follows:
\begin{itemize}
    \item[\scalebox{0.6}{$\blacksquare$}] We introduce FrequencyLowCut pooling, ensuring aliasing-free down-sampling within CNNs.
    \item[\scalebox{0.6}{$\blacksquare$}] Through extensive experiments with various datasets and architectures, we show empirically that FLC pooling prevents single step AT from catastrophic overfitting, while this is not the case for other recently published improved pooling operations (e.g.~\cite{zhang2019making}).
    \item[\scalebox{0.6}{$\blacksquare$}] FLC pooling is substantially faster, around five times, and easier to integrate than previous AT or defence methods. It provides a hyperparameter-free plug and play module for increased model robustness.
\end{itemize}

\subsection{Related Work}

\noindent{\textbf{Adversarial Attacks.}}
Adversarial attacks reveal CNNs vulnerabilities to intentional pixel perturbations which are crafted either having access to the full model (so-called white-box attacks) \cite{harnesssing,deepfool,intriguing,deepfool,c_and_w,pgd,decoupling} or only having access to the model's prediction on given input images (so-called back-box attacks) \cite{andriushchenko2020square,cheng2018query}. 
The Fast Gradient Sign Method~\cite{harnesssing}, FGSM, is an efficient single step white box attack. More effective methods use multiple optimization steps, e.g. as in the white-box Projected Gradient Descent (PGD) \cite{pgd} or in black-box attacks such as Squares \cite{andriushchenko2020square}.  AutoAttack \cite{auto_attack} is an ensemble of different attacks including an adaptive version of PGD and is widely used to benchmark adversarial robustness because of its strong performance \cite{robust_bench}. 
In relation to image down-sampling, \cite{xiao2017wolf} and \cite{lohn2020downscaling} demonstrate steganography-based attacks on the pre-processing pipeline of CNNs. 

\noindent{\textbf{Adversarial Training.}}
Some adversarial attacks are directly proposed with a dedicated defence \cite{harnesssing,decoupling}. Beyond these attack-specific defences, there are many methods for more general adversarial training (AT) schemes.
These typically add an additional loss term which accounts for possible perturbations \cite{robustness_github,trades} or introduces additional training data~\cite{carmon2019unlabeled,sehwag2021improving}. Both are combined for example in \cite{mart}, while \cite{gowal2021uncovering} use data augmentation which is typically combined with weight averaging \cite{rebuffi2021fixing}.
A widely used source for additional training data is \textit{ddpm} \cite{gowal2021improving,rade2021helperbased,rebuffi2021fixing}, which contains one million extra samples for CIFAR-10 and is generated with the model proposed by \cite{ho2020denoising}.
\cite{gowal2021improving} receive an additional boost in robustness by adding specifically generated images while \cite{rade2021helperbased} add wrongly labeled data to the training-set. 
RobustBench \cite{robust_bench} gives an overview and evaluation of a variety of models w.r.t.~their adversarial robustness and the additional data used.

\noindent A common drawback of all AT methods is the vast increase in computation needed to train networks: large amounts of additional adversarial samples and slower convergence due to the harder learning problem typically increase the training time by a factor between seven and fifteen \cite{pgd,mart,wu2020adversarial,trades}.    

\noindent{\textbf{Catastrophic Overfitting.}}
AT with single step FGSM is a simple approach to achieve basic adversarial robustness \cite{chen2021robust,rice2020overfitting}. Unfortunately, the robustness of this approach against stronger attacks like PGD is starting to drop again after a certain amount of training epochs. \cite{wong2020fast} called this phenomenon {\it catastrophic overfitting}. They concluded that one step adversarial attacks tend to overfit to the chosen adversarial perturbation magnitude (given by $\epsilon$) but fail to be robust against multi-step attacks like PGD. \cite{rice2020overfitting} introduced early stopping as a countermeasure. After each training epoch, the model is evaluated on a small portion of the dataset with a multi-step attack, which again increases the computation time. As soon as the accuracy drops compared with a hand selected threshold the model training is stopped. \cite{kim2020understanding} and  \cite{stutz2021relating} showed that the observed overfitting is related to the flatness of the loss landscape. They introduced a method to compute the \textit{optimal} perturbation length $\epsilon'$ for each image and do single step FGSM training with this optimal perturbation length to prevent catastrophic overfitting.  \cite{andriushchenko2020understanding} showed that catastrophic overfitting not only occurs in deep neural networks but can also be present in single-layer convolutional neural networks. They propose a new kind of regularization, called GradAlign to improve FGSM perturbations and flatten the loss landscape to prevent catastrophic overfitting.

\noindent{\textbf{Anti-Aliasing.}}
The problem of aliasing effects in the context of CNN-based neural networks has already been addressed from various angles in literature: \cite{zhang2019making} improve the shift-invariance of CNNs using anti-aliasing filters implemented as convolutions. \cite{zou2020delving} further improve shift invariance by using learned instead of predefined blurring filters. \cite{li2021wavecnet} rely on the low frequency components of wavelets during pooling operations to reduce aliasing and increase the robustness against common image corruptions. In \cite{hossain2021antialiasing} a depth adaptive blurring filter before pooling as well as an anti-aliasing activation function are used. Anti-aliasing is also relevant in the context of image generation. \cite{karras2021aliasfree} propose to use blurring filters to remove aliases during image generation in generative adversarial networks (GANs) while \cite{durall2020watch} and \cite{jung2020spectral} employ additional loss terms in the frequency space to address aliasing.
 In \cite{grabinski2022aliasing}, we empirically showed via a proposed aliasing measure that adversarially robust models exhibit less aliasing in their down-sampling layers than non-robust models. Based on this motivation, we here propose an aliasing-free down-sampling operation that avoids catastrophic overfitting.

\section{Preliminaries}
\subsection{Adversarial Training}
In general, AT can be formalized as an optimization problem given by a {\it min-max} formulation: 
\begin{equation}
    \min_{\theta} \max_{\delta \in\Delta} L(x+\delta, y; \theta)\,,
\end{equation}
where we seek to optimize network weights $\theta$ such that they minimize the loss $L$ between inputs $x$ and labels $y$ under attacks $\delta$. The maximization over $\delta$ can thereby be efficiently performed using the Fast Gradient Sign Method ({FGSM}), which takes one big step defined by $\epsilon$ into the direction of the gradient \cite{harnesssing}: 
\begin{equation}
    x' = x+ \epsilon\cdot \mathrm{sign}(\nabla_x L(\theta, x, y))\, .
\end{equation}
Specific values of the perturbation size $\epsilon$ are usually set to be fractions of eight-bit encodings of the image color channels. A popular choice on the CIFAR-10~\cite{cifar} dataset is $\epsilon={\frac{8}{255}}$ which can be motivated by the human color perception \cite{Gonzalez}. 
The Projected Gradient Descent method, PGD, works similar to FGSM but instead of taking one big step in the direction of the gradient with step size $\epsilon$, it iteratively optimizes the adversarial example with a smaller, defined step size $\alpha$. Random restarts further increase its effectiveness. The final attack is clipped to the maximal step size of $\epsilon$.
\begin{equation}
    x'_{N+1} = \mathrm{Clip}_{X,\epsilon}\{x'_N + \alpha \cdot \mathrm{sign}(\nabla_x L(\theta, x, y))\}
\end{equation}
PGD is one of the strongest attacks, due to its variability in step size and its random restarts. Yet, its applicability for AT is limited as it requires a relatively long optimization time for every example. Additionally, PGD is dependent on several hyperparameters, which makes it even less attractive for training in practice. In contrast, FGSM is fast and straight-forward to implement. Yet, models that use FGSM for AT tend to overfit on FGSM attacks and are not robust to other attacks such as PGD, i.e.~they suffer from catastrophic overfitting~\cite{wong2020fast}.  

\subsection{Down-sampling in CNNs}
Independent of their actual network topology, CNNs essentially perform a series of stacked convolutions and non-linearities. Using a vast amount of learnable convolution filters, CNNs are capable of extracting local texture information from all intermediate representations (input data and feature maps). To be able to abstract from this localized spatial information and to learn higher order relations of parts, objects and entire scenes, CNNs apply down-sampling operations to implement a spatial pyramid representation over the network layers. 

\begin{figure*}[ht]
 \scriptsize
\begin{center}
\begin{tabular}{@{}c@{}c@{}}
\begin{minipage}{.48\textwidth}
      \includegraphics[width=\linewidth]{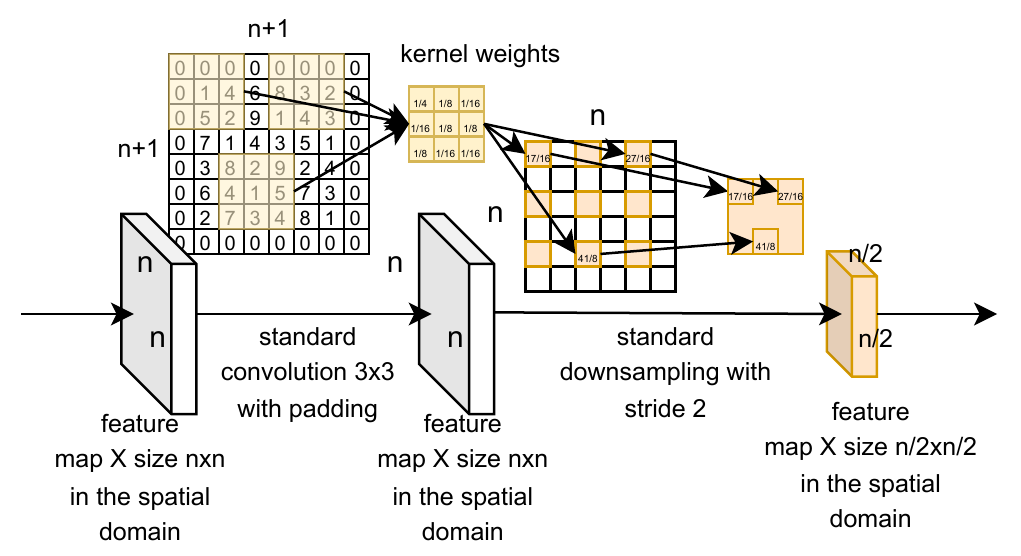}
     \end{minipage}  
&
  \begin{minipage}{.48\textwidth}
      \includegraphics[width=\linewidth]{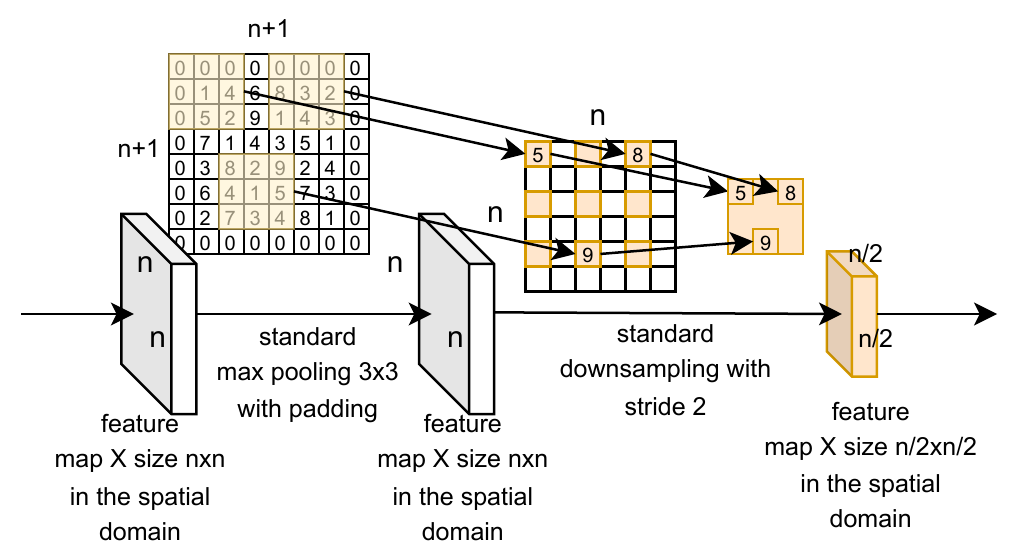}
     \end{minipage}  
 \end{tabular}
 \\
 \caption[]{Standard down-sampling operations used in CNNs. Left: down-sampling via convolution with stride two. First the feature map is padded and the actual convolution is executed. The stride defines the step-size of the kernel. Hence, for stride two, the kernel is moved two spatial units. In practice, this down-sampling is often implemented by a standard convolution with stride one and then discarding every second point in every spatial dimension. Right: down-sampling via MaxPooling. Here the max value for each spatial window location is chosen and the striding is implemented accordingly.}
 \label{fig:standard_down}
 \end{center}
 \end{figure*}

This down-sampling is typically performed via a convolution with stride greater than one or by so-called pooling layers (see Fig. \ref{fig:standard_down}). The most common pooling layers are AveragePooling and MaxPooling.  
All of these operations are highly sensitive to small shifts or noise in the layer input \cite{li2021wavecnet,chaman2021truly,zhang2019making}.

\begin{figure}[t]
	\begin{center}
	\includegraphics[width=\linewidth]{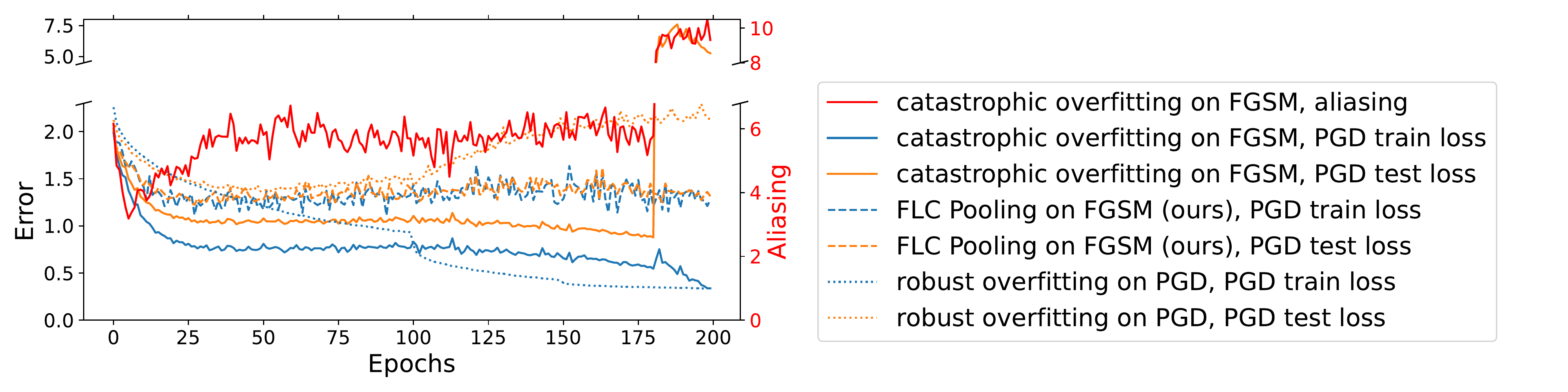}
	\caption[]{Examples of AT facing catastrophic overfitting and its relationship to aliasing as well as robust overfitting and our FLC pooling. While FGSM training is prone to catastrophic overfitting, PGD training takes much longer and is also prone to robust overfitting. Our method, FLC pooling, is able to train with the fast FGSM training while preventing catastrophic overfitting.}
	\label{fig:relationship}
	\end{center}
\end{figure}

\begin{figure}[t]
	\begin{center}
	\includegraphics[width=0.9\linewidth]{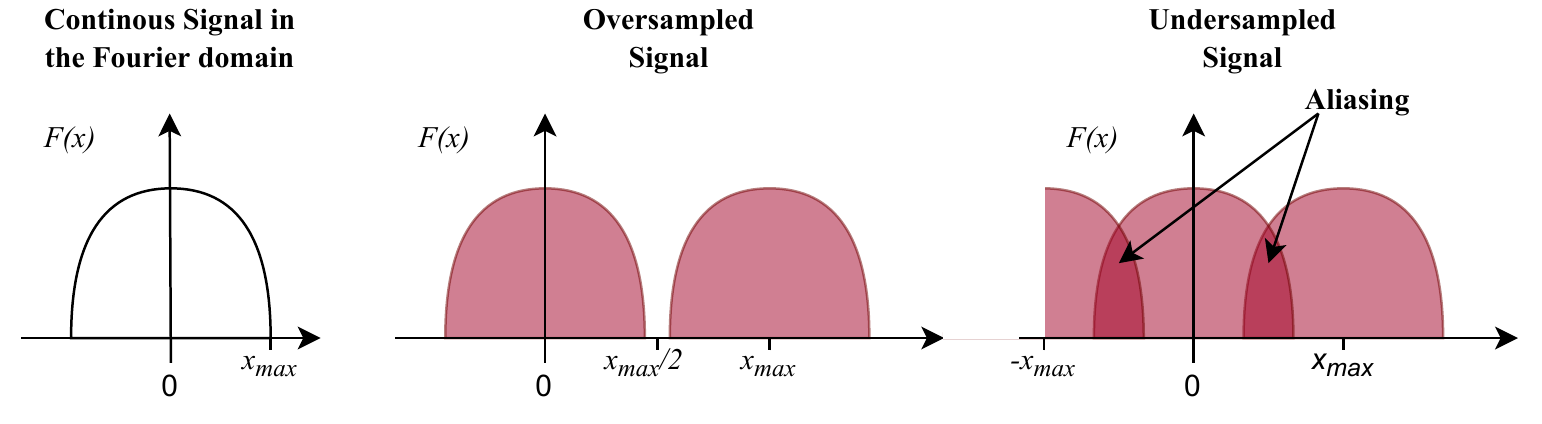}
	\caption[]{Aliasing is apparent in the frequency domain. Left: The frequency spectrum of a 1D signal with maximal frequency $x_{\mathrm{max}}$. After down-sampling, replica of the signal appear at a distance proportional to the sampling rate. Center: The spectrum after sampling with a sufficiently large sampling rate. Right: The spectrum after under-sampling with aliases due to overlapping replica.}
	\label{fig:aliasing_theory}
	\end{center}
\end{figure}

\noindent{\textbf{Aliasing.}}
Common CNNs sub-sample their intermediate feature maps to aggregate spatial information and increase the invariance of the network. However, no aliasing prevention is incorporated in current sub-sampling methods. Concretely, sub-sampling with too low sampling rates will cause pathological overlaps in the frequency spectra (Fig. \ref{fig:aliasing_theory}). They arise as soon as the sampling rate is below the double bandwidth of the signal \cite{shannon} and cause ambiguities: high frequency components can not be clearly distinguished from low frequency components. As a result, CNNs might misconceive local uncorrelated image perturbations as global manipulations. \cite{grabinski2022aliasing} showed that aliasing in CNNs strongly coincides with the robustness of the model. Based on this finding, one can hypothesize that models that overfit to high frequencies in the data tend to be less robust. This thought is also in line with the widely discussed texture bias \cite{geirhos2018imagenet}. To substantiate this hypothesis in the context of adversarial robustness, we investigate and empirically show in Figure \ref{fig:relationship} that catastrophic overfitting coincides with increased aliasing during FGSM AT. Based on this observation, we expect networks that sample without aliasing to be better behaved in AT FGSM settings. The FrequencyLowCut pooling, which we propose, trivially fulfills this property.

\section{FrequencyLowCut Pooling}
Several previous approaches such as \cite{zhang2019making,zou2020delving} reduce high frequencies in features maps before pooling to avoid aliasing artifacts. They do so by classical blurring operations in the spatial domain. While those methods reduce aliasing, they can not entirely remove it due to sampling theoretic considerations in theory and limited filter sizes in practice (see Appendix \ref{sec:free_down} or \cite{Gonzalez} for details). 
We aim to perfectly remove aliases in CNNs' down-sampling operations without adding additional hyperparameters. Therefore, we directly address the down-sampling operation in the frequency domain, where we can sample according to the Nyquist rate, i.e.~remove all frequencies above $\frac{\mathrm{sampling rate}}{2}$~and thus discard aliases. 
In practice, the proposed down-sampling operation first performs a Discrete Fourier Transform (DFT) of the feature maps $f$. Feature maps with height $M$ and width $N$ to be down-sampled are then represented as
\begin{equation}
\label{fft}
    F(k,l)= \frac{1}{MN} \sum_{m=0}^{M-1} \sum_{n=0}^{N-1} f(m,n)e^{-2\pi j\left(\frac{k}{M}m+\frac{l}{N}n\right)}\,.
\end{equation}
In the resulting frequency space representation $F$ (eq.~\eqref{fft}), all frequencies $k,l$, with $|k|$ or $|l|>\frac{\mathrm{sampling rate}}{2}$ have to be set to $0$ before down-sampling. CNNs commonly down-sample with a factor of two, i.e.~sampling rate $=\frac{1}{2}$. Down-sampling thus corresponds to finding  
$
F_d(k,l)=F(k,l) \,, \forall\,\mathrm{frequencies}\,k,l\,  \mathrm{with}\, |k|,|l|<\frac{1}{4}.
$
Practically, the DFT$(f)$ returns an array $\mathrm{F}$ of complex numbers with size $K\times L$ = $M\times N$, where the frequency $k,l=0$ is stored in the upper left corner 
and the highest frequency is in the center. 
\begin{figure}[t]
	\begin{center}
	\includegraphics[width=0.7\linewidth]{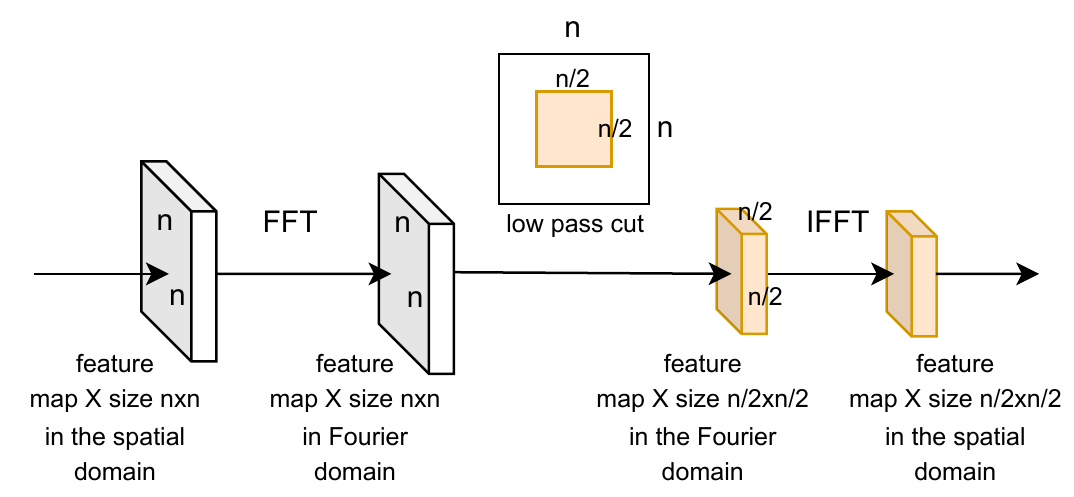}
	\caption[]{FrequencyLowCut pooling, the proposed, guaranteed alias-free pooling operation. We first transform feature maps into frequency space via FFT, then crop the low frequency components. The result is transformed back into the spatial domain. This corresponds to a sinc-filtered and down-sampled feature map and is fed into the next convolutional layer.}
	\label{fig:low_cut_pool}
	\end{center}
\end{figure}
We thus shift the low frequency components into the center of the array via FFT-shift to get $\mathrm{F}_s$  and crop the frequencies below the Nyquist frequency as $\mathrm{F}_{sd}=\mathrm{F}_s[K^\prime:3K^\prime, L^\prime:3L^\prime]$ for $K^\prime=\frac{K}{4}$ and $L^\prime=\frac{L}{4}$, for all samples in a batch and all channels in the feature map. After the inverse FFT-shift, we obtain array $\mathrm{F}_d$ with size $[\hat{K},\hat{L}]=[\frac{K}{2},\frac{L}{2}]$, containing exactly all frequencies below the Nyquist frequency $F_d$, which we can backtransform to the spatial domain via inverse DFT for the spatial indices $\hat{m}=0\dots \frac{M}{2}$ and $\hat{n}=0\dots \frac{N}{2}$. 
\begin{equation}
    f_d(\hat{m}, \hat{n}) =  \frac{1}{\hat{K}\hat{L}} \sum_{k=0}^{\hat{K}-1} \sum_{l=0}^{\hat{L}-1} F_d(k,l)e^{2\pi j\left(\frac{\hat{m}}{\hat{K}}k+\frac{\hat{n}}{\hat{L}}l\right)}.
\end{equation}
We thus receive the aliasing-free down-sampled feature map $f_d$ with size $[\frac{M}{2},\frac{N}{2}]$.

Fig. \ref{fig:low_cut_pool} shows this procedure in detail. In the spatial domain, this operation would amount to convolving the feature map with an infinitely large (non-bandlimited) $\mathrm{sinc}(m)=\frac{\sin(m)}{m}$ filter, which can not be implemented in practice.

\section{Experiments}

\subsection{Native Robustness of FLC pooling}
We evaluate our proposed FLC pooling in a standard training scheme with Preact-ResNet-18 (PRN-18) architectures on CIFAR-10 (see Appendix \ref{ref:train_schedule} for details). Table \ref{tab:native} shows that both the decrease in clean accuracy as well as the increase in robustness are marginal compared to the baseline models. We argue that these results are in line with our hypothesis that the removal of aliasing artifacts alone will not lead to enhanced robustness and we need to combine correct down-sampling with AT to compensate for the persisting problems induced by the very high dimensional decision spaces in CNNs.    
\setlength{\tabcolsep}{0.8em}
\begin{table}[t]
\tiny
\caption[]{Clean training of Preact-ResNet-18 (PRN-18) architectures on CIFAR-10. We compare clean and robust accuracy against FGSM~\cite{harnesssing} with $L_{\inf}$, $\epsilon=\frac{8}{255}$, PGD~\cite{pgd} with $L_{\inf}$, $\epsilon=\frac{1}{255}$ as well as $L_2$ with $\epsilon=0.5$ (20 iterations) and common corruptions (CC)~\cite{hendrycks2019using} (mean over all corruptions and severities).}
\scriptsize
\begin{center}
\begin{tabular}{l|cccc|c}
\toprule
Method                                     & Clean       &   \makecell{FGSM  \\ $\epsilon=\frac{8}{255}$}          &\makecell{PGD $L_{\inf}$ \\ $\epsilon=\frac{1}{255}$ } & \makecell{PGD $L_{2}$ \\ $\epsilon=0.5$ } & CC
\\ \midrule

Baseline          & \textbf{95.08 }     & 34.08  &  7.15  & 6.68  &74.38 \\
FLC Pooling        &   94.66   &  \textbf{34.65 }&  \textbf{10.00} & \textbf{11.27} &\textbf{74.70} 
\\ \bottomrule
\end{tabular}
\label{tab:native}
\end{center}
\end{table}
\setlength{\tabcolsep}{0.5em}
\begin{table}[t]
\caption[]{FGSM AT of PRN-18 and Wide-ResNet-28-10 (WRN-28-10) architectures on CIFAR-10. Comparison of clean and robust accuracy (high is better) against PGD \cite{pgd} and AutoAttack \cite{auto_attack} on the full dataset with $L_{\inf}$ with $\epsilon=8/255$ and $L_2$ with $\epsilon=0.5$. FGSM test accuracies indicate catastrophic overfitting on the AT data, hence this column is set to gray.}
\scriptsize
\begin{center}
\begin{tabular}{l|c>{\color{gray}\arraybackslash}ccccc}
\toprule
Method                                     & Clean       &   \makecell{FGSM  \\ $\epsilon=\frac{8}{255}$}          & \makecell{PGD $L_{\inf}$ \\ $\epsilon=\frac{8}{255}$} & \makecell{AA  $L_{\inf}$ \\ $\epsilon=\frac{8}{255}$}  & \makecell{AA  $L_2$ \\ $\epsilon=0.5$ }   &  \makecell{AA $L_{\inf} $ \\ $\epsilon=\frac{1}{255}$  }  \\ \midrule
\midrule 
\multicolumn{7}{c}{ \textbf{Preact-ResNet-18}}\\
\hline 
Baseline: FGSM training                 & \textbf{90.81}   &           90.37        & 0.16            & 0.00        & 0.01  & 53.10  \\ 
Baseline \& early stopping & 82.88   &    61.71                 & 11.82          &    3.76   &   17.44 &72.95\\ 
BlurPooling \cite{zhang2019making}             &  86.24 &78.36  & 1.33 & 0.06 &1.96 & 66.88\\ 
Adaptive BlurPooling \cite{zou2020delving}              & 90.35& 77.39 &0.23 & 0.00 & 0.07 & 39.00\\ 
Wavelet Pooling \cite{li2020wavelet}           & 85.02 & 64.16 &12.13& 5.92 &19.65 &10.08\\ 
FLC Pooling (ours)             & 84.81 &  58.25 &\textbf{38.41}  &  \textbf{36.69} &\textbf{55.58} &\textbf{80.63}\\ 
\hline 
\midrule
\multicolumn{7}{c}{\textbf{WRN-28-10}}\\
\hline 
Baseline: FGSM training                      &        86.67    & 83.64           &        1.64     &   0.09   & 1.47 & 59.39\\ 
Baseline    \& early stopping &             82.29  &     56.36    &      31.26      &  28.54 & 46.03 &  76.87\\ 
Blurpooling \cite{zhang2019making}                    &         91.40 &      89.44 &   0.22  & 0.00 & 0.00 & 38.45 \\ 
Adaptive BlurPooling \cite{zou2020delving}                  & 91.10& 89.76  & 0.00 & 0.00 & 0.00 & 7.42\\ 
Wavelet Pooling \cite{li2020wavelet}                  &\textbf{92.19}& 90.85  & 0.00 &    0.00 & 0.00 & 10.08\\ 
FLC Pooling (ours)       &      84.93 & 53.81
 & \textbf{39.48 } &  \textbf{38.37} &\textbf{52.89} & \textbf{80.27}\\ \bottomrule
\end{tabular}
\label{tab:cifar10}
\end{center}
\end{table}
\setlength{\tabcolsep}{0.5em}
\begin{table}[t]
\caption[]{FGSM AT on CINIC-10 for PRN-18 architectures. We compare clean and robust accuracy (higher is better) against PGD \cite{pgd} as well as AutoAttack \cite{auto_attack} on the full dataset with $L_{\inf}$ with $\epsilon=8/255$ and $L_2$ with $\epsilon=0.5$. FGSM test accuracies indicate catastrophic overfitting on the AT data, hence this column is set to gray.}
\scriptsize
\begin{center}
\begin{tabular}{l|c>{\color{gray}\arraybackslash}ccccc}
\toprule
Method                                     & Clean         &   \makecell{FGSM  \\ $\epsilon=\frac{8}{255}$}              & \makecell{PGD $L_{\inf}$ \\ $\epsilon=\frac{8}{255}$} & \makecell{AA  $L_{\inf}$ \\ $\epsilon=\frac{8}{255}$}  & \makecell{AA  $L_2$ \\ $\epsilon=0.5$ }   &  \makecell{AA $L_{\inf} $ \\ $\epsilon=\frac{1}{255}$  }  \\ \midrule
Baseline                  &     87.46   &    58.83           &         1.31   &       0.12  & 1.55 &55.21 \\ 
Baseline  \& early stopping &   82.79    &    42.58            &         27.55        &   30.76  & 50.28 & \textbf{79.88} \\ 
Blurpooling \cite{zhang2019making}                 &      87.13   &   54.16  &   1.29        & 0.20 & 4.68 &70.56\\ 
Adaptive BlurPooling \cite{zou2020delving}       & \textbf{90.21} & 52.27  &0.05    & 0.00&  0.01 &40.96 \\ 
aWavelet Pooling \cite{li2020wavelet}              & 88.81&  64.16   & 1.76&   0.12 &3.38 &66.61\\
 FLC Pooling (ours)    &  82.56 &   38.39  &\textbf{36.28} & \textbf{49.61} &  \textbf{60.51} & 78.50\\ \bottomrule
\end{tabular}
\label{tab:cinic10}
\end{center}
\end{table}
\subsection{FLC pooling for FGSM training}
In the following series of experiments we apply simple FGSM AT with $\epsilon=\frac{8}{255}$ on different architectures and evaluate the resulting robustness with different pooling methods.   
We compare the models in terms of their clean, FGSM, PGD and AutoAttack accuracy, where the FGSM attack is run with $\epsilon = 8/255$,  PGD with 50 iterations and 10 random restarts and $\epsilon = 8/255$ and $\alpha = 2/255$. For AutoAttack, we evaluate the standard $L_{\inf}$ norm with $\epsilon = 8/255$ and a smaller $\epsilon$ of $1/255$, as AutoAttack is almost too strong to be imperceptible to humans \cite{lorenz2022is}. Additionally, we evaluate AutoAttack with $L_2$ norm and $\epsilon=0.5$.

\noindent{\textbf{CIFAR-10.}} Table \ref{tab:cifar10} shows the evaluation of a PRN-18 as well as a Wide-ResNet-28-10 (WRN-28-10) on CIFAR-10 \cite{cifar}. For both network architectures, we observe that our proposed FLC pooling is the only method that is able to prevent catastrophic overfitting. All other pooling methods heavily overfit on the FGSM training data, achieving high robustness towards FGSM attacks, but fail to generalize towards PGD or AutoAttack. Our hyper-parameter free approach also outperforms early stopping methods which are additionally suffering from the difficulty that one has to manually choose a suitable threshold in order to maintain the best model robustness. 

\noindent{\textbf{CINIC-10.}} Table \ref{tab:cinic10} shows similar results on CINIC-10 \cite{darlow2018cinic}. Our model exhibits no catastrophic overfitting, while previous pooling methods do. It should be noted that CINIC-10 is not officially reported by AutoAttack. This might explain why the accuracies under AutoAttack are higher on CINIC-10 than on CIFAR-10. We assume that AutoAttack is optimized for CIFAR-10 and CIFAR-100 and therefore less strong on CINIC-10.
\setlength{\tabcolsep}{0.5em}
\begin{table}[t]
\tiny
\caption[]{FGSM AT on CIFAR-100 for PRN-18 architectures. We compare clean and robust accuracy (higher is better) against PGD \cite{pgd} and AutoAttack \cite{auto_attack} on the full dataset with $L_{\inf}$ with $\epsilon=8/255$ and $L_2$ with $\epsilon=0.5$. FGSM test accuracies indicate robustness to training data, so this column is set to gray. Here, none of the models overfit, while FLC pooling still yields best overall robustness.}
\label{tab:cifar100}
\scriptsize
\begin{center}
\begin{tabular}{l|c>{\color{gray}\arraybackslash}ccccc}
\toprule
Method                                     & Clean          &   \makecell{FGSM  \\ $\epsilon=\frac{8}{255}$}             & \makecell{PGD $L_{\inf}$ \\ $\epsilon=\frac{8}{255}$} & \makecell{AA  $L_{\inf}$ \\ $\epsilon=\frac{8}{255}$}  & \makecell{AA  $L_2$ \\ $\epsilon=0.5$ }   &  \makecell{AA $L_{\inf} $ \\ $\epsilon=\frac{1}{255}$  }  \\ \midrule
Baseline                      &               51.92          &   23.25 &          15.41      &   11.13 &25.67 &44.53\\ 
Baseline    \& early stopping &            52.09           &   23.34 &      15.51   &      10.88  & 25.78 & 44.61\\ 
Blurpooling \cite{zhang2019making}               & 52.68& 23.40 &16.81   & 12.43 &26.79 & 45.68\\
Adaptive BlurPooling \cite{zou2020delving}                    & 52.08 &  9.77&18.68 &  6.05  &  11.32 &21.04\\ 
Wavelet Pooling \cite{li2020wavelet}                  & \textbf{55.08} & 25.70& 18.36 & 13.76  &  \textbf{27.51} &47.52 \\ 
FLC Pooling (ours)                   &  54.66 & 26.82  &\textbf{19.83 } & \textbf{15.40} &   26.30 & \textbf{47.83}\\ \bottomrule
\end{tabular}
\end{center}
\end{table}

\noindent{\textbf{CIFAR-100.}} Table \ref{tab:cifar100} shows the results on CIFAR-100 \cite{cifar}, using the same experimental setup as for CIFAR-10 in Table \ref{tab:cifar10}. Due to the higher complexity of CIFAR-100, with ten times more classes than CIFAR-10, AT tends to suffer from catastrophic overfitting much later (in terms of epochs) in the training process. Therefore we trained the Baseline model for 300 epochs. While the gap towards the robustness of other methods is decreasing with the amount of catastrophic overfitting, our method still outperforms other pooling approaches in most cases - especially on strong attacks.

\setlength{\tabcolsep}{1.2em}
\noindent{\textbf{ImageNet.}} 
\begin{table}
\scriptsize
  \caption{Comparison of ResNet-50 models clean and robust accuracy against AutoAttack \cite{auto_attack} on ImageNet. We compare against models reported on RobustBench \cite{robust_bench}.}
    \centering
    \begin{tabular}{l|cc}
    \toprule
    Method & Clean & \makecell{PGD $L_{\inf}$\\
    $\epsilon = \frac{4}{255}$} \\
    \midrule
    Standard \cite{robust_bench} & 76.52 & 0.00 \\
    \midrule
         FGSM \& FLC Pooling (ours)  & 63.52 & 27.29 \\
         Wong et al., 2020 \cite{wong2020fast} & 55.62 & 26.24\\
         \midrule
         Robustness lib, 2019 \cite{robustness_github} & 62.56 & 29.22\\
         Salman et al., 2020 \cite{salman2020adversarially} & 64.02 & 34.96\\
         \bottomrule
    \end{tabular}
    \label{tab:imagenet}
\end{table}
Table \ref{tab:imagenet} evaluates our FLC Pooling on ImageNet.
We compare against results reported on RobustBench \cite{robust_bench}, with emphasis on the model by \cite{wong2020fast} which also uses fast FGSM training.
The clean accuracy of our model using FLC pooling is about 8\% better than the one reached by \cite{wong2020fast}, with a 1\% improvement in robust accuracy. All other models are trained with more time consuming methods like PGD (more details can be found in the Appendix~\ref{sec:imagenet_time}).\\

\noindent{\textbf{Analysis.}}
The presented experiments on several datasets and architectures show that baseline FGSM training, as well as other pooling methods, strongly overfit on the adversarial data and do not generalize their robustness towards other attacks. We also show that our FLC pooling sufficiently prevents catastrophic overfitting and is able to generalize robustness over different networks, datasets, and attack sizes in terms of different $\epsilon$-values.

\begin{figure}
 \scriptsize
\begin{center}
\begin{adjustbox}{width=0.95\textwidth}
\begin{tabular}{c@{\,}cc@{\,}cc@{\,}c}
 & Example Image & \makecell{Spectrum of the\\ Example Image} & \makecell{Spectrum Difference \\ to the Original \\ Image} & \makecell{Spatial Difference \\ to the Original Image \\ mean over 100 images} & \makecell{Spectrum Difference \\ to the Original Image \\ mean over 100 images} \\
 \rotatebox[origin=c]{90}{Original Image} &
 \begin{minipage}{.24\textwidth}
      \includegraphics[width=\linewidth]{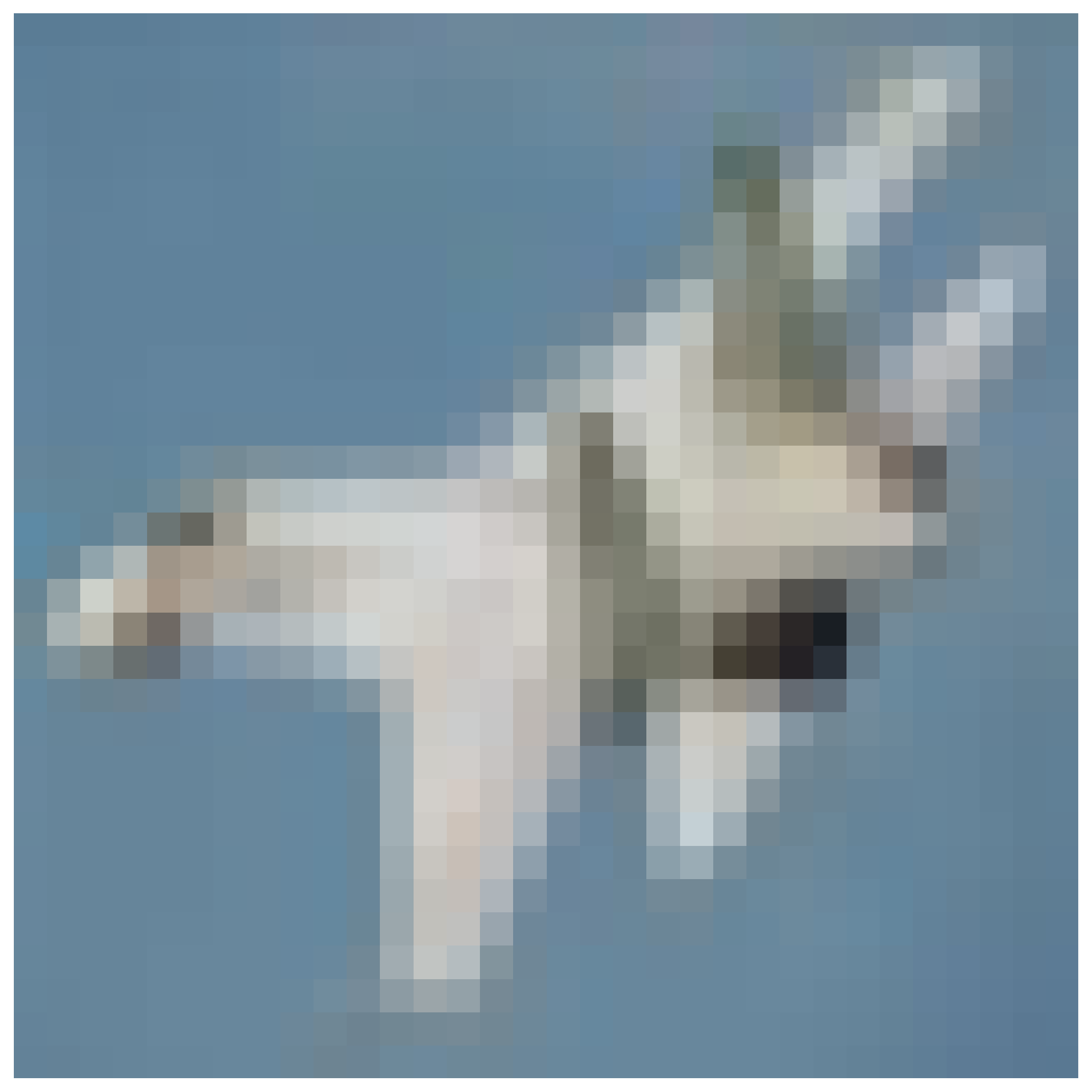}
    \end{minipage} &
 \begin{minipage}{.24\textwidth}
      \includegraphics[width=\linewidth]{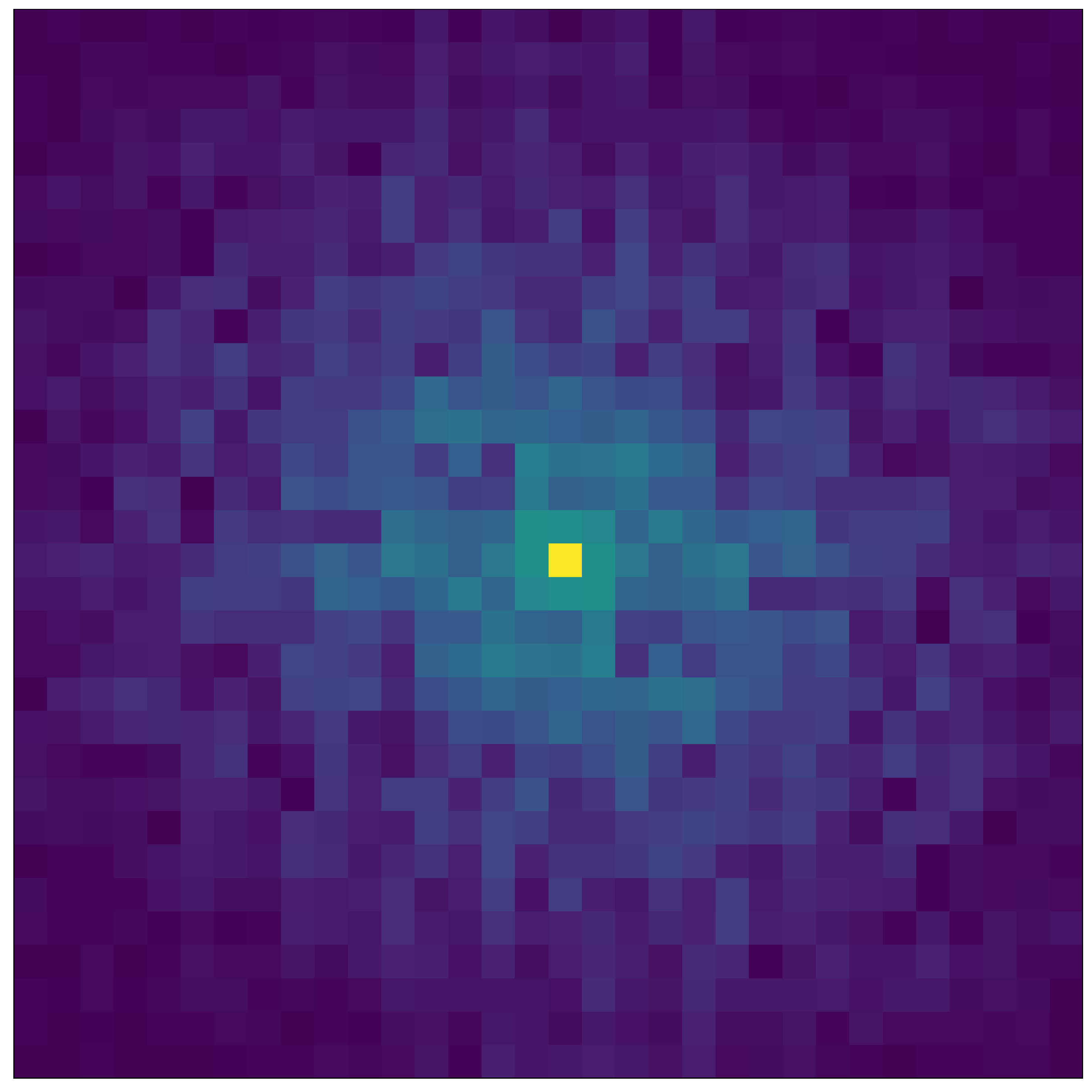}
    \end{minipage} &
\begin{minipage}{.24\textwidth}
      \includegraphics[width=\linewidth]{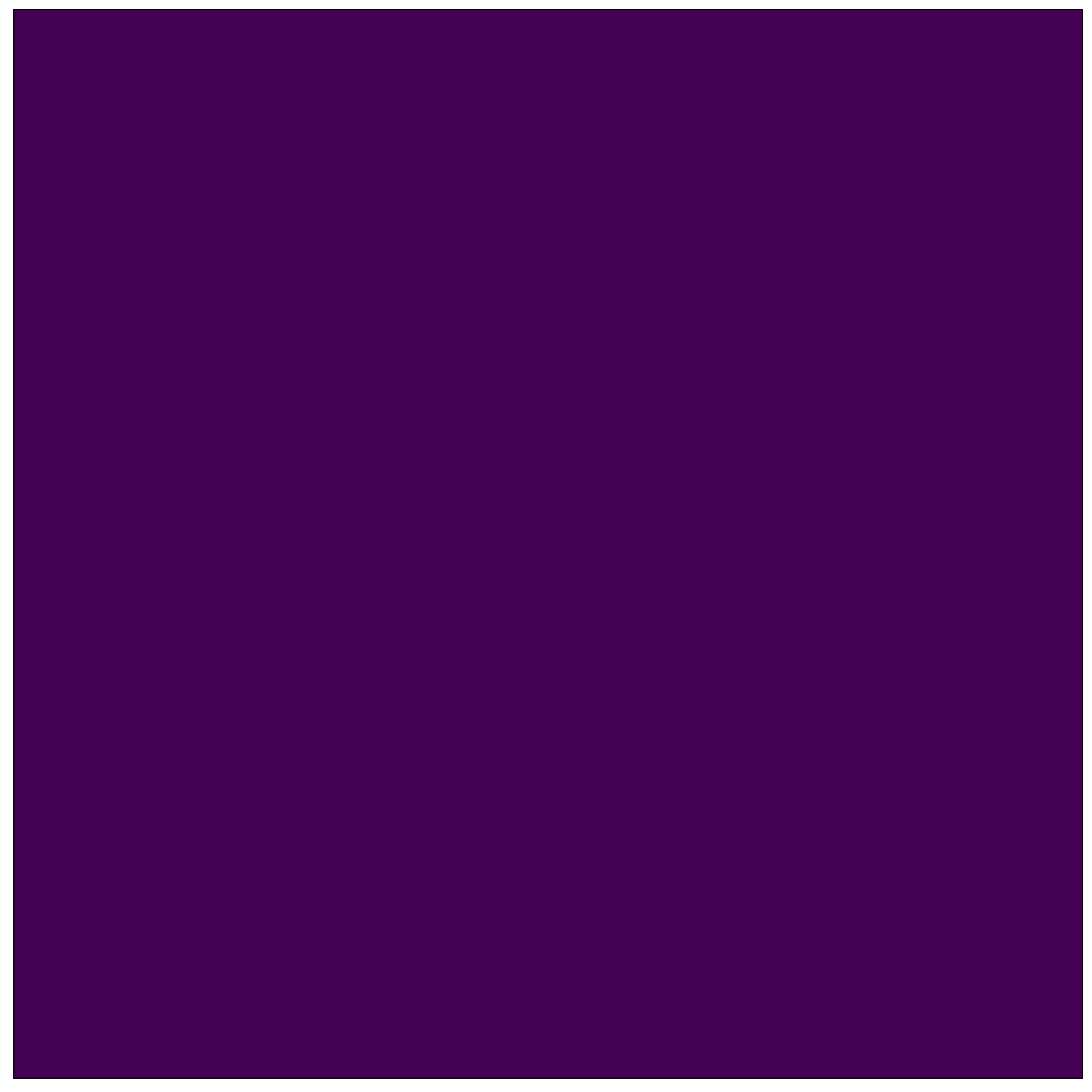}
    \end{minipage} &
\begin{minipage}{.24\textwidth}
      \includegraphics[width=\linewidth]{auto_attack_spectrum_zero_10_Airplane.pdf}
     \end{minipage} &
 \begin{minipage}{.24\textwidth}
       \includegraphics[width=\linewidth]{auto_attack_spectrum_zero_10_Airplane.pdf}
     \end{minipage}  \\ 
  \rotatebox[origin=c]{90}{\makecell{Baseline \\ Perturbation}} &   
 \begin{minipage}{.24\textwidth}
      \includegraphics[width=\linewidth]{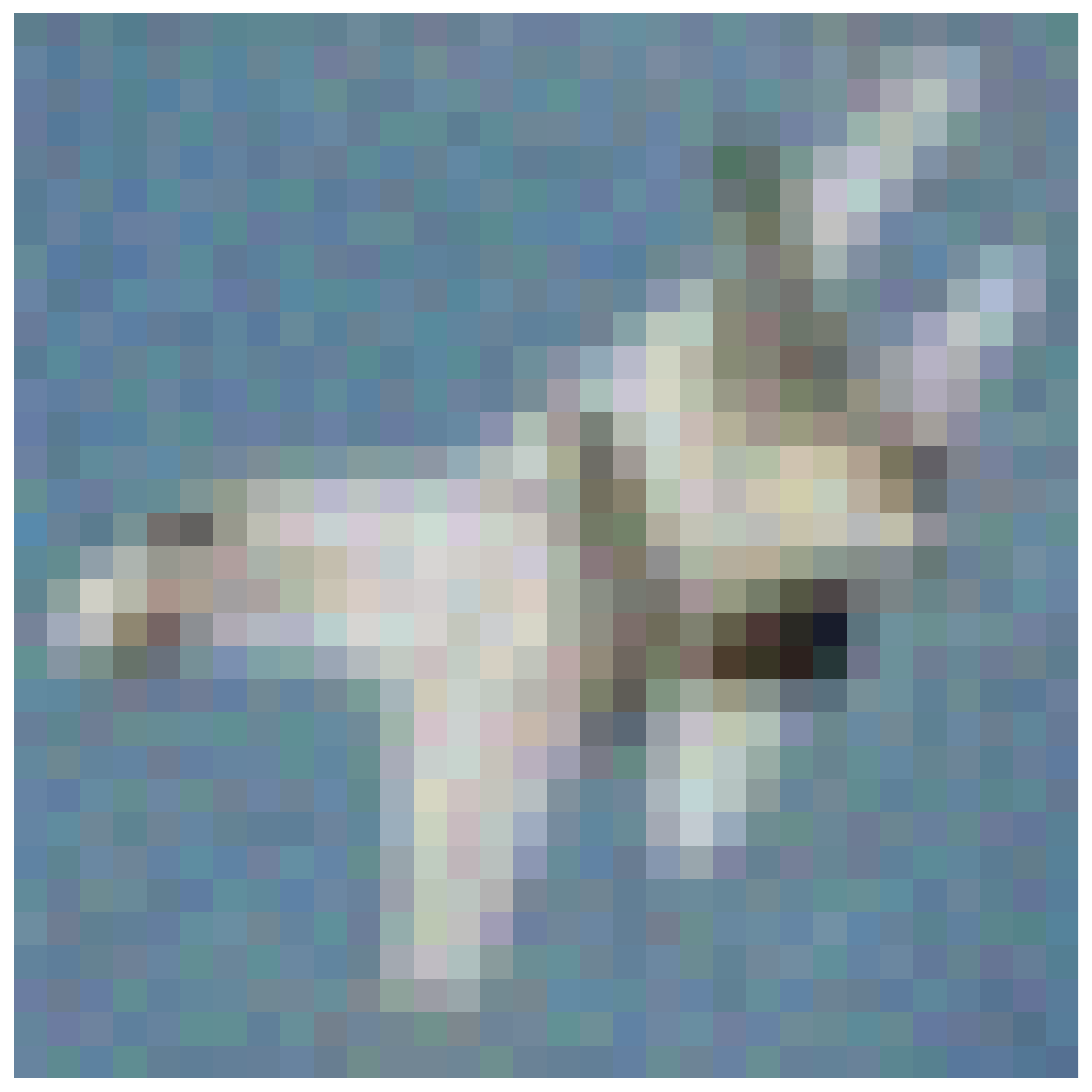}
    \end{minipage} &
 \begin{minipage}{.24\textwidth}
      \includegraphics[width=\linewidth]{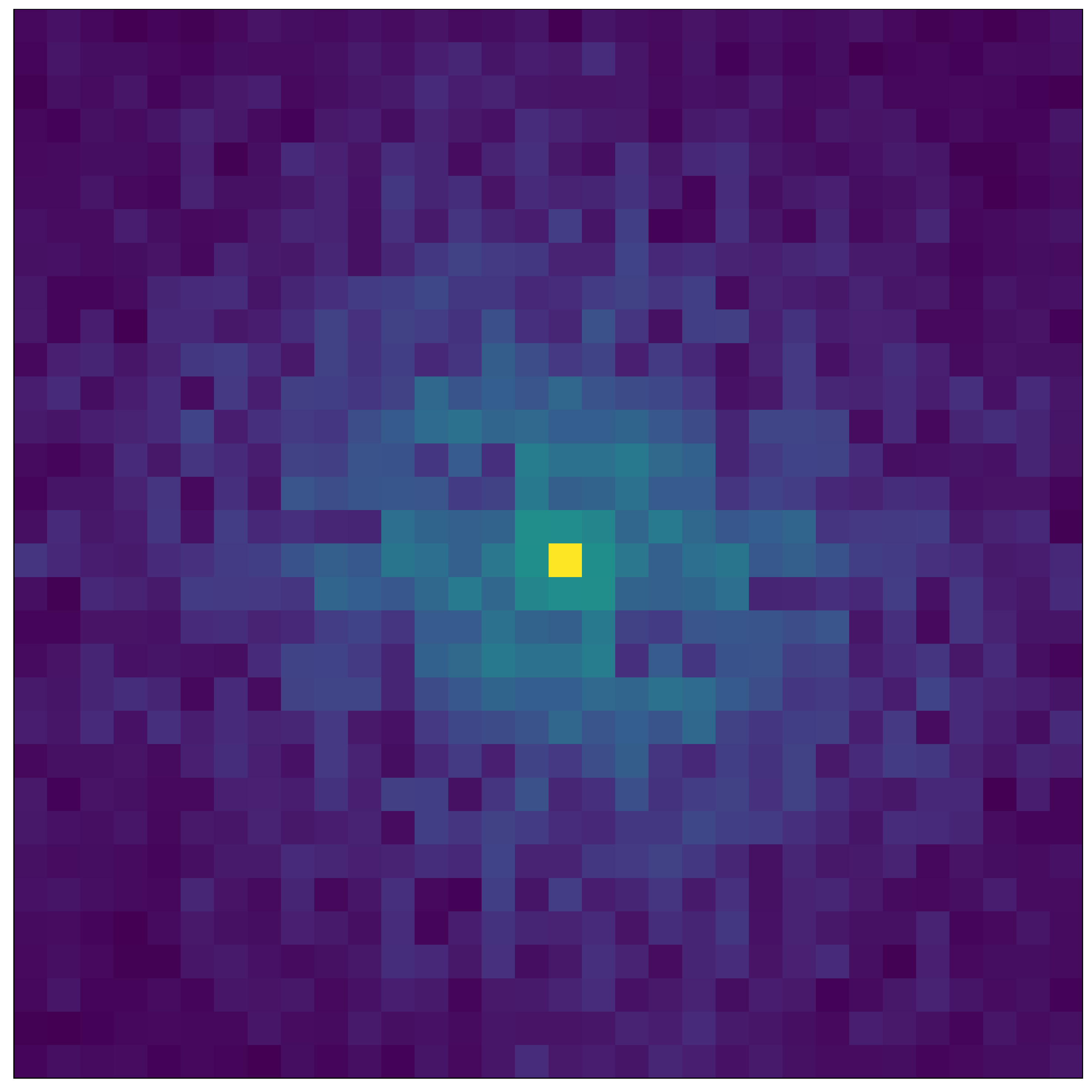}
    \end{minipage} &
\begin{minipage}{.24\textwidth}
      \includegraphics[width=\linewidth]{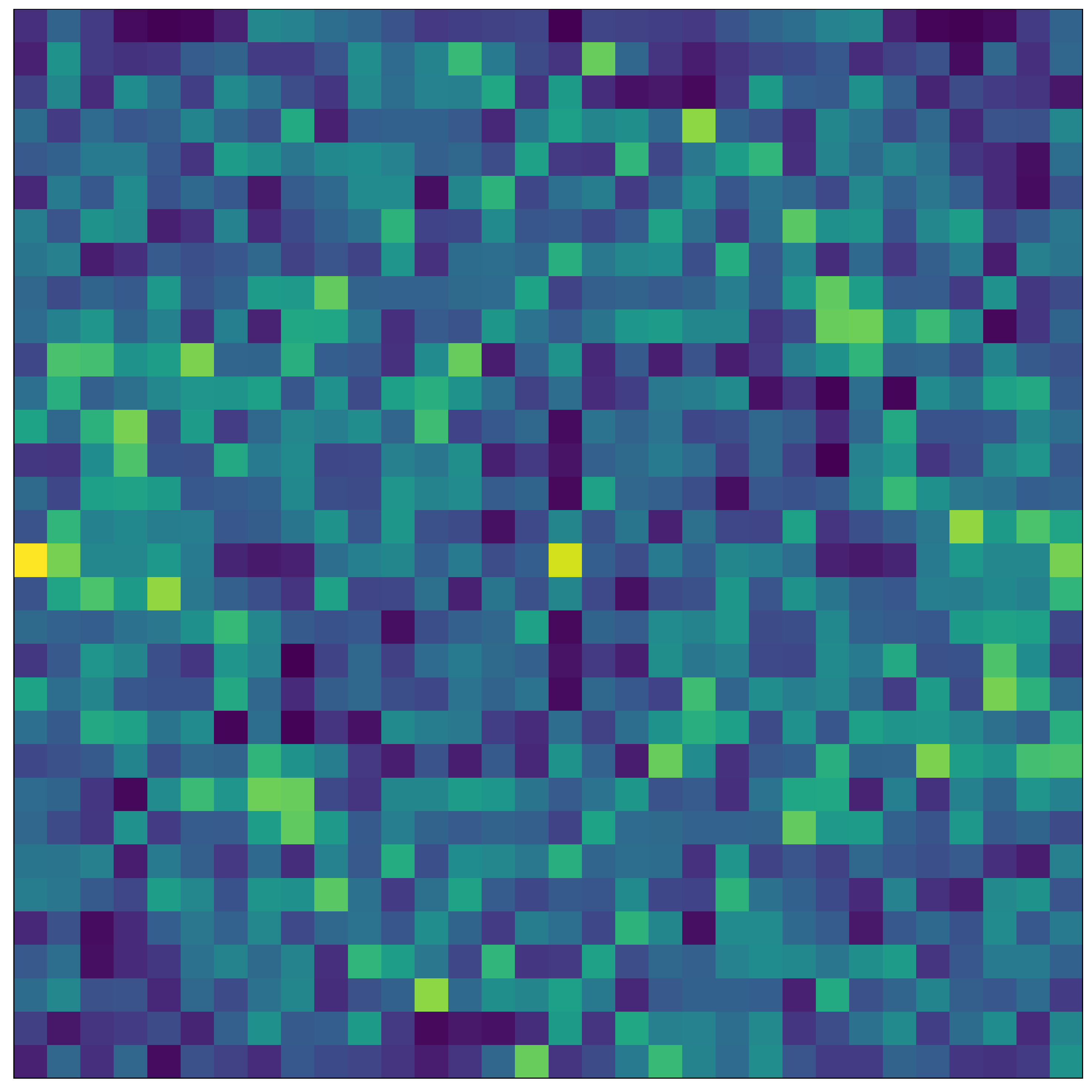}
    \end{minipage} &
\begin{minipage}{.24\textwidth}
      \includegraphics[width=\linewidth]{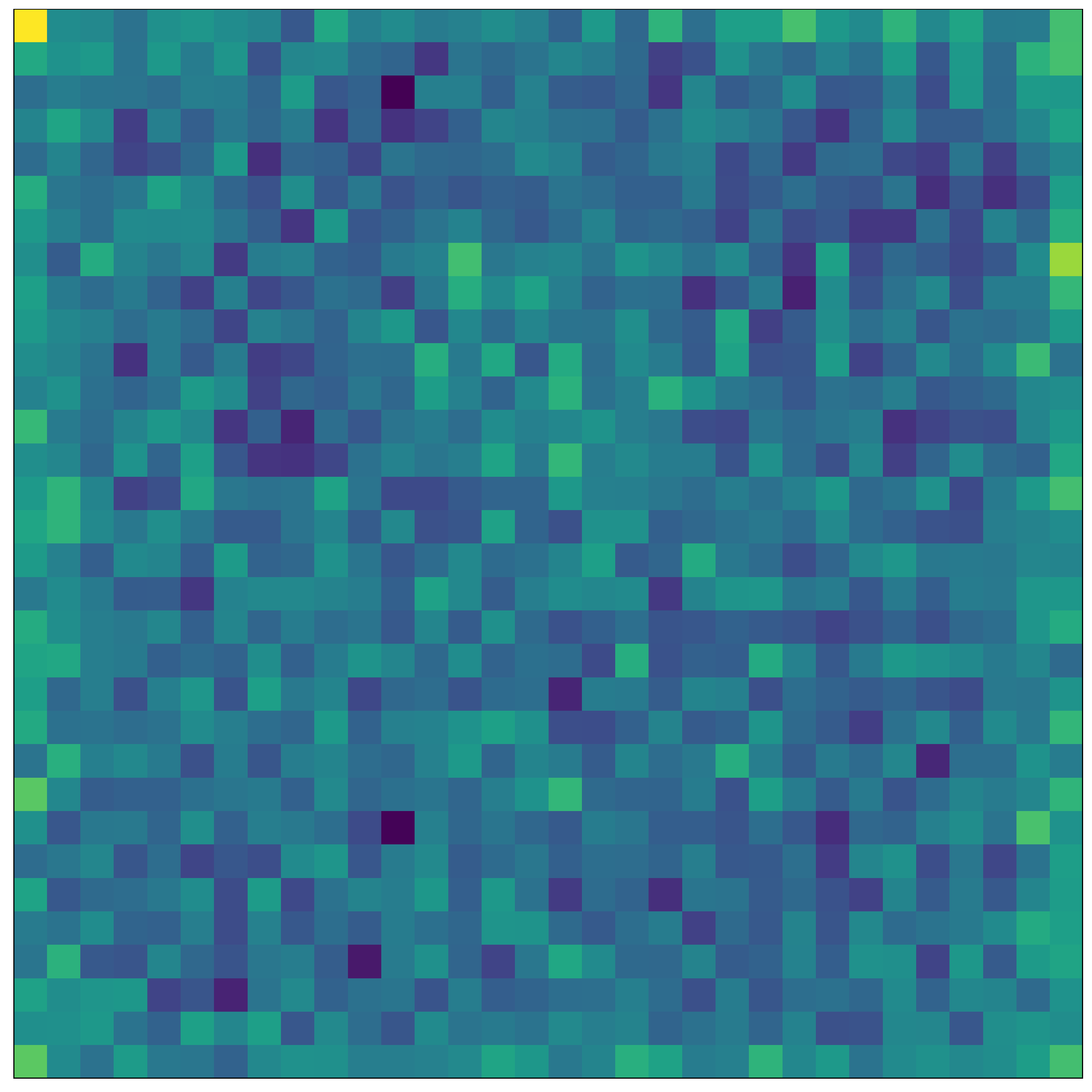}
     \end{minipage} &
 \begin{minipage}{.24\textwidth}
       \includegraphics[width=\linewidth]{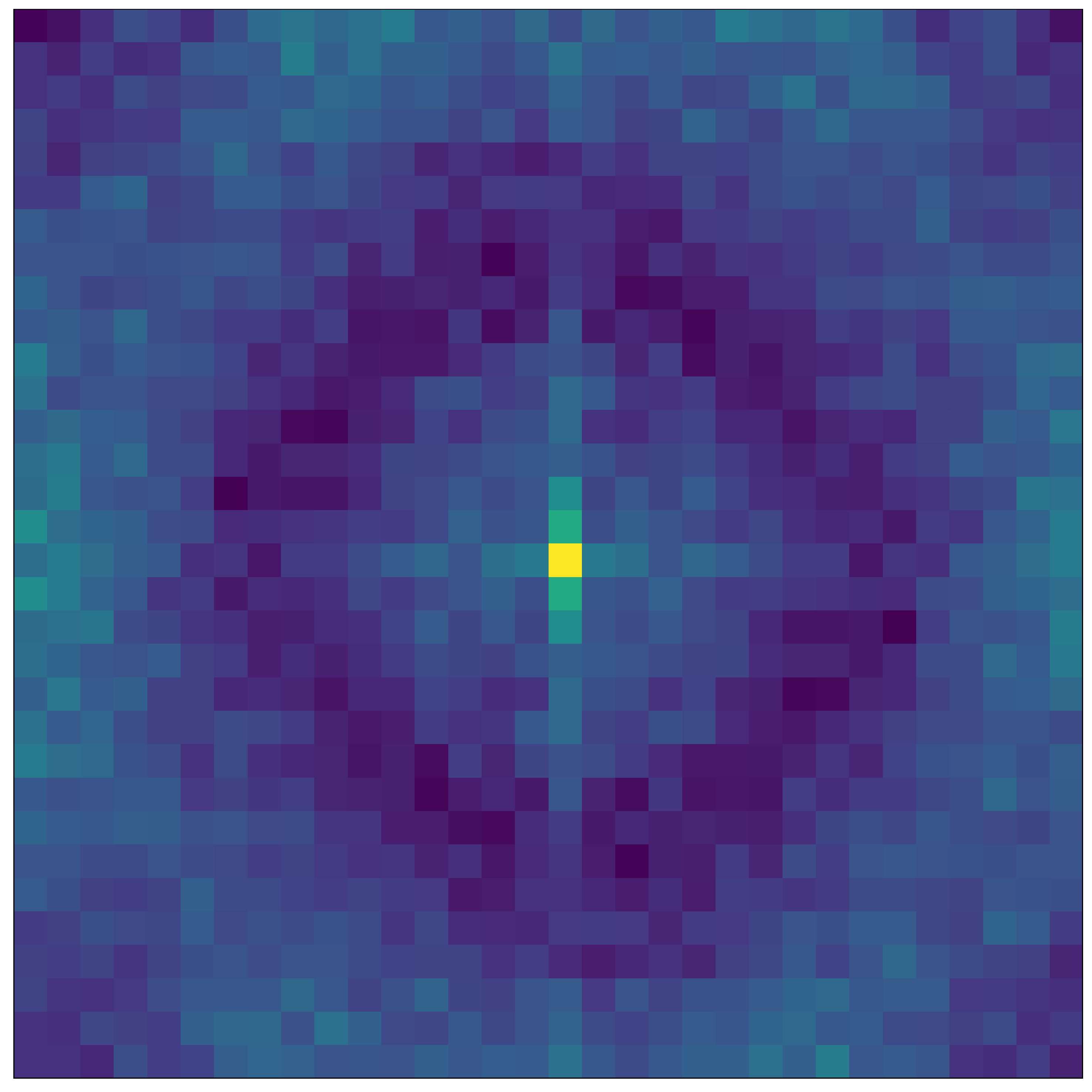}
     \end{minipage}  \\
  \rotatebox[origin=c]{90}{\makecell{FLC Pooling \\ Perturbations}} &   
 \begin{minipage}{.24\textwidth}
      \includegraphics[width=\linewidth]{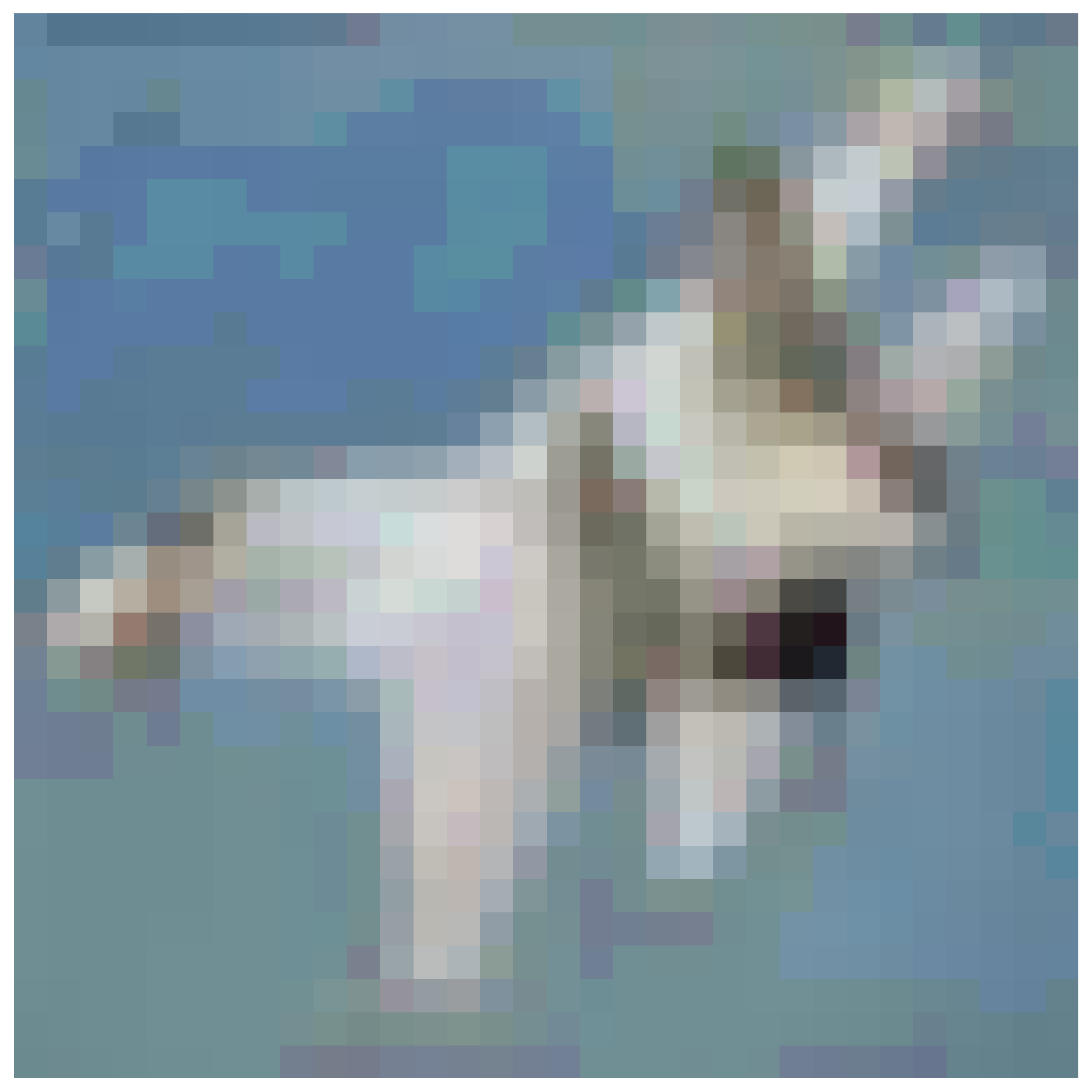}
    \end{minipage} &
 \begin{minipage}{.24\textwidth}
      \includegraphics[width=\linewidth]{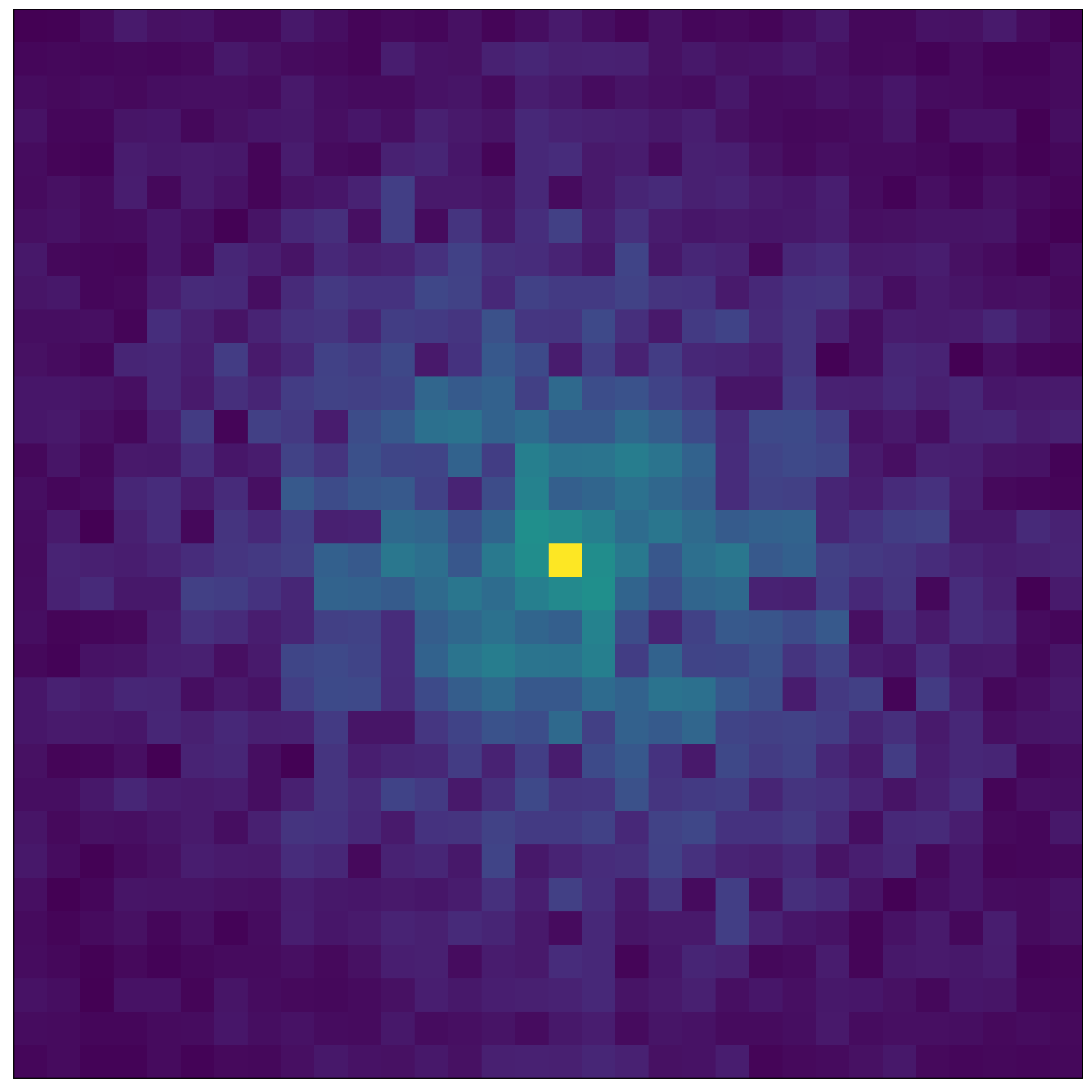}
    \end{minipage} &
\begin{minipage}{.24\textwidth}
      \includegraphics[width=\linewidth]{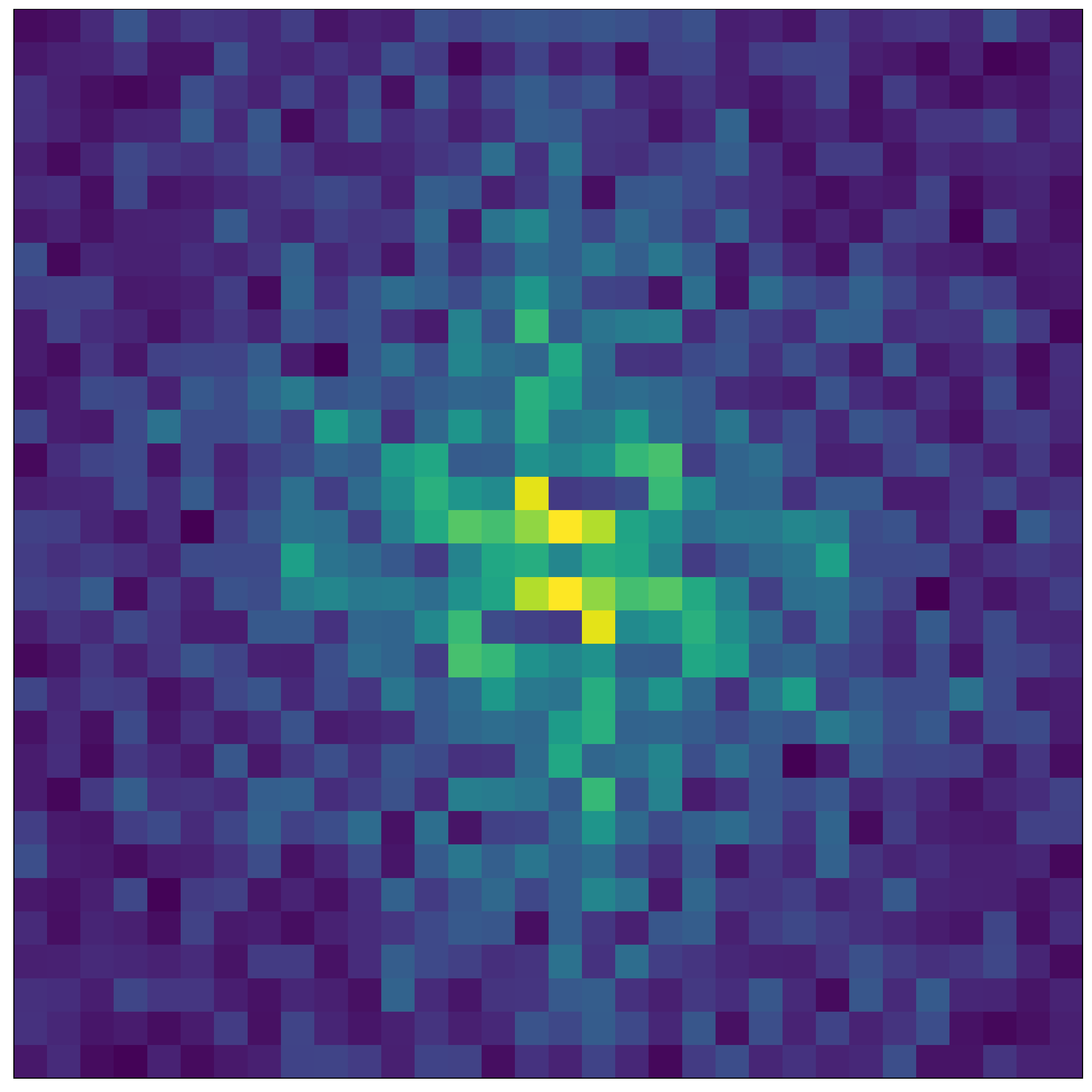}
    \end{minipage} &
\begin{minipage}{.24\textwidth}
      \includegraphics[width=\linewidth]{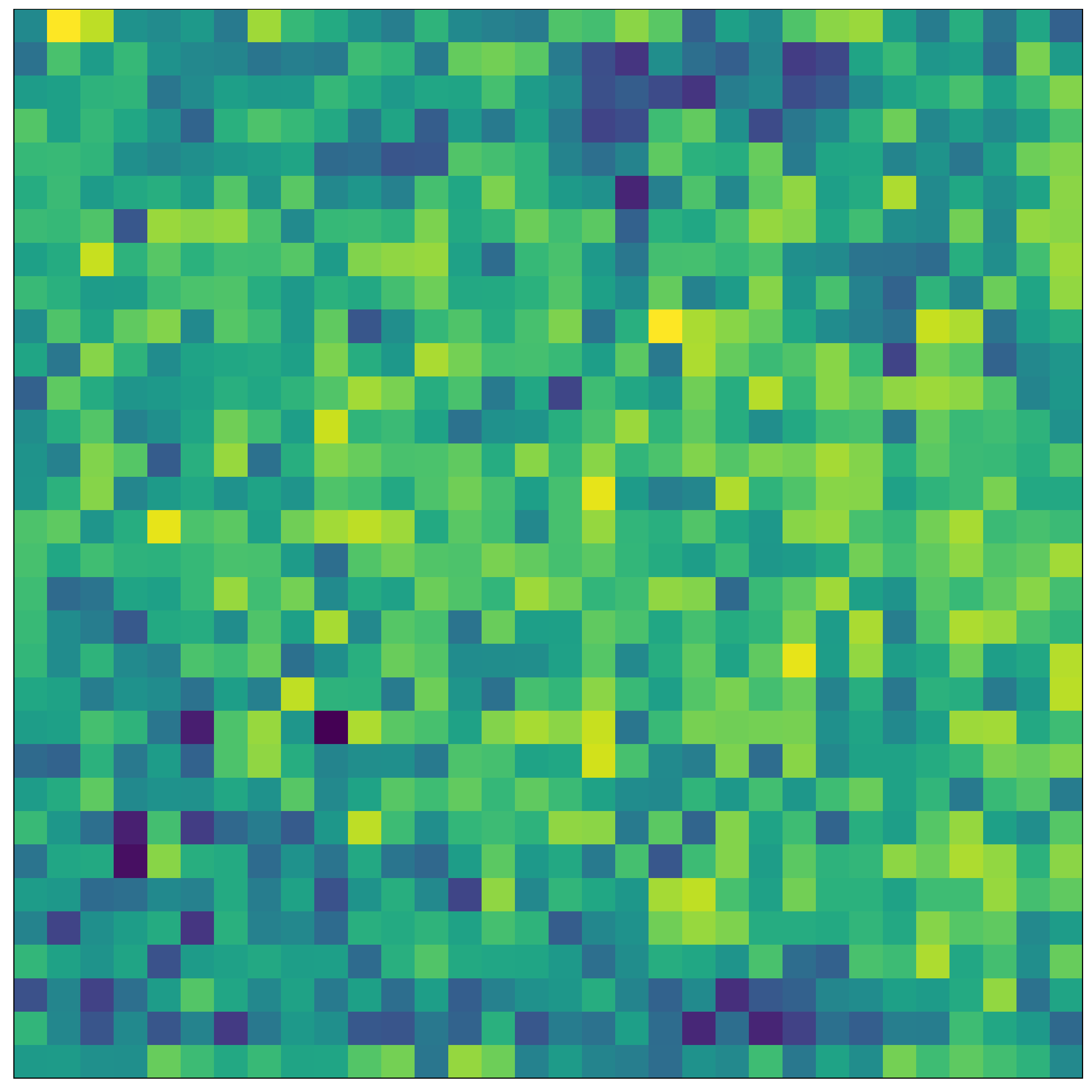}
     \end{minipage} &
 \begin{minipage}{.24\textwidth}
       \includegraphics[width=\linewidth]{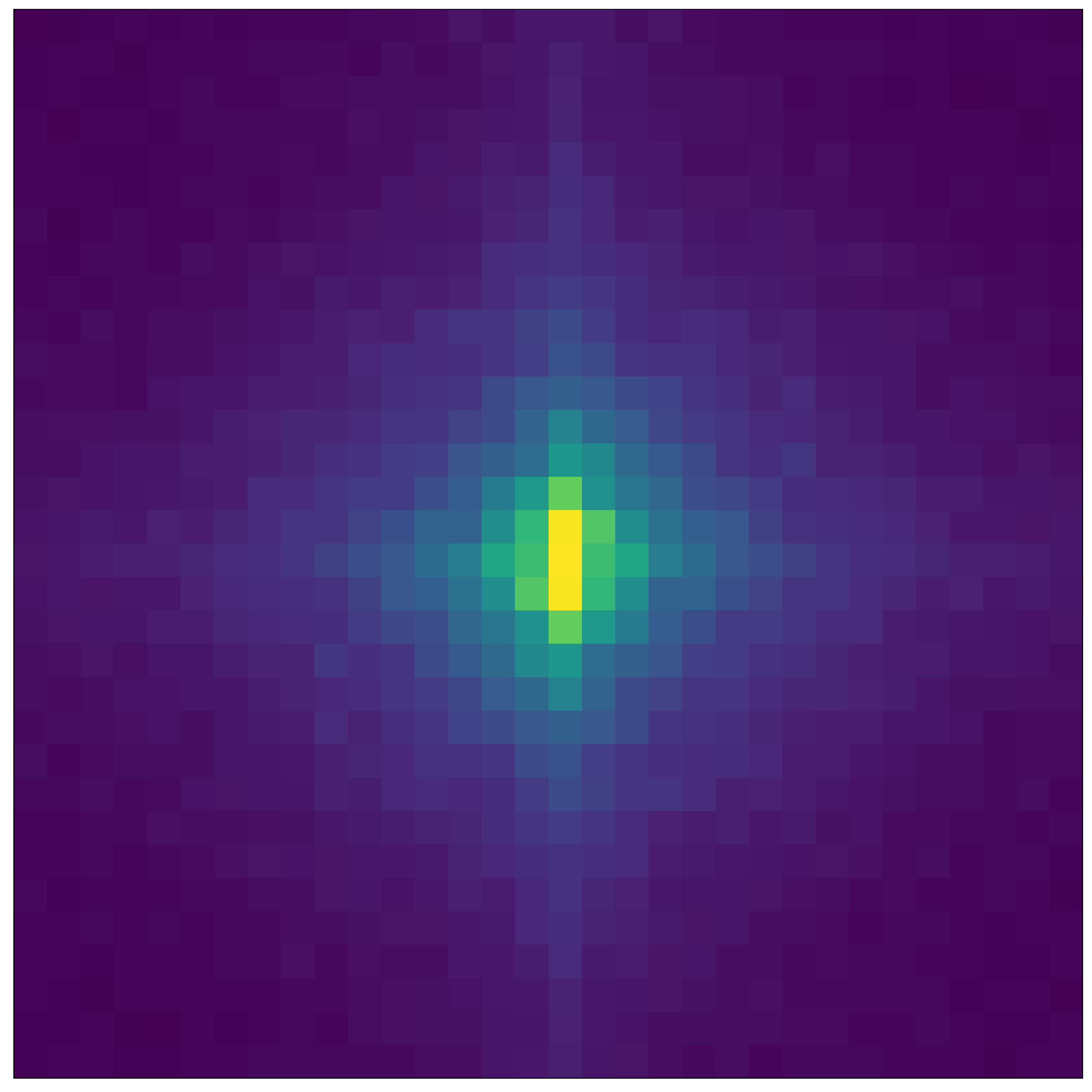}
     \end{minipage}   \\
 \end{tabular}
 \end{adjustbox}
 \caption[]{Spatial and spectral differences of adversarial perturbations created by AutoAttack with $\epsilon=\frac{8}{255}$ on the baseline model as well as our FLC Pooling. On the left side for one specific example of an airplane and on the right side the average difference over 100 images.}
 \label{fig:attack_images}
 \end{center}
 \end{figure}

\noindent{\textbf{Attack Structures.}}
In Figure \ref{fig:attack_images}, we visualize AutoAttack adversarial attacks.Perturbations created for the baseline trained with FGSM differ substantially from those created for FLC pooling trained with FGSM. While perturbations for the baseline model exhibit high frequency structures, attacks to FLC pooling rather affect the global image structure.

\subsection{Training Efficiency}
Most AT approaches use adversarial image perturbations during training \cite{harnesssing,pgd} \cite{wong2020fast}. Thereby the time and memory needed depend highly on the specific attack used to generate the perturbations. Multi-step attacks like PGD \cite{pgd} require substantially more time than single step attacks like FGSM \cite{wong2020fast}. TRADES \cite{trades} incorporates different loss functions to account for a good trade-off between clean and robust accuracy. With our FLC pooling, we provide a simple and fast method for more robust models. Therefore we compare our method with state-of-the-art training schedules in terms of time needed per epoch when trained in their most basic form. Table \ref{tab:time_report} shows that FGSM training is fastest. However, FGSM with early stopping is not able to maintain high robustness against AutoAttack \cite{auto_attack} due to catastrophic overfitting. PGD training can establish robustness against AutoAttack. It relies on the same training procedure as FGSM but uses expensive multi-step perturbations and thereby increases the computation time by over a factor of four (4.23). For Adversarial Weight Perturbations (AWP) the training time per epoch is over six times (6.57), for TRADES by eight times (8.04) higher. Our FLC pooling increases the training only by a factor of 1.26 while achieving a good clean and robust accuracy. 
When adding additional data like the \textit{ddpm} dataset to the training as it is done in all leading RobustBench \cite{robust_bench} models, the training time is increased by a factor of twenty. The {\it ddpm} dataset incorporates one million extra samples, which is over sixteen times more than the original CIFAR-10 dataset. We report our training times for ImageNet in Appendix \ref{sec:imagenet_time} to show that FLC pooling is scalable in terms of practical runtime.
\setlength{\tabcolsep}{0.5em}
\begin{table}[t]
\small
\caption[]{Runtime of AT in seconds per epoch over 200 epochs and a batch size of 512 trained with a PRN-18 for training on the original CIFAR-10 dataset without additional data. Experiments are executed on one Nvidia Tesla V100. Evaluation for clean and robust accuracy, higher is better, on AutoAttack \cite{auto_attack} with our trained models. The models reported by the original authors may have different numbers due to different hyperparameter selection. 
The top row reports the baseline without AT.}
\scriptsize
\begin{center}
\begin{adjustbox}{width=0.9\textwidth}
\begin{tabular}{l|c|cc}
\toprule
Method                      &     Seconds per epoch (avg)     & Clean Acc & AA Acc    \\ \midrule
Baseline & 14.6 $\pm$ 0.1 & 95.08 & 0.00 \\ \midrule
FGSM    \& early stopping \cite{wong2020fast} &     \textbf{27.3 $\pm$ 0.1 }  & 82.88       & 11.82\\ 
FGSM    \& FLC Pooling (Ours) &     34.5 $\pm$ 0.1   & \textbf{84.81} & 38.41  \\ 
PGD \cite{pgd} &       115.4 $\pm$ 0.2 & 83.11 & 40.35 \\ 
Robustness lib \cite{robustness_github} &  117 $\pm$  19.0 &76.37 & 32.10\\
AWP \cite{wu2020adversarial}&  179.4 $\pm$  0.4 &82.61 & \textbf{49.43}\\ 
MART \cite{Wang2020Improving} & 180.4 $\pm$ 0.8 &55.49 & 8.63 \\
TRADES \cite{trades} &     219.4 $\pm$ 0.5   & 81.49 & 46.91 \\ \bottomrule
\end{tabular}
\end{adjustbox}
\label{tab:time_report}
\end{center}
\end{table}
\subsection{Black Box Attacks} 
PGD and AutoAttack are intrinsically related to FGSM. Therefore, to allow for a clean evaluation of the model robustness without bias towards the training scheme, we also evaluate black box attacks.
Squares \cite{andriushchenko2020square}, which is also part of the AutoAttack pipeline, adds perturbations in the form of squares onto the image until the label flips. Besides Squares, we evaluate two transferred perturbations. The first perturbation set is constructed through the baseline network which is not robust at all. The second set is constructed from the baseline network which is trained with FGSM and early stopping. We evaluate against different PRN-18 and WRN-28-10 models on CIFAR-10 as well as PRN-18 models on CIFAR-100 provided by RobustBench \cite{robust_bench}. Note that all networks marked with * are models which rely on additional data sources such as \textit{ddmp}~\cite{ho2020denoising}. Other RobustBench models like \cite{gowal2021improving} rely on training data that is not available anymore such that fair comparison is currently not possible. Arguably, we always expect models to further improve as training data is added.
\setlength{\tabcolsep}{0.5em}
\begin{table}[t]
\caption[]{Robustness against black box attacks on PRN-18 and WRN-28-10 models with CIFAR-10. First against Squares \cite{andriushchenko2020square} with $\epsilon=1/255$ and then against perturbations which were created on the baseline network, meaning transferred perturbations (TP), and the baseline model including early stopping (TPE). As well as the accuracy under common corruptions (CC).}
\scriptsize
\begin{center}
\begin{tabular}{l|lccc|c}
\toprule
Model                   & Clean & Squares & TP &TPE & CC \\ \midrule
\midrule 
\multicolumn{6}{c}{ \textbf{Preact-ResNet-18}}\\
\hline 
Baseline & \textbf{90.81}& 78.04& 0.00 &69.33 & 71.81   \\
FGSM    \& early stopping & 82.88 & 77.58   &   77.67       &        3.76    &         71.80      \\ 
FGSM \& FLC Pooling (ours)          & 84.81 & \textbf{81.40} &    \textbf{83.64}       &     \textbf{80.49}      & \textbf{76.15}       \\ 
Andriushchenko and Flammarion, 2020 \cite{andriushchenko2020understanding}   &       79.84       & 76.78 & 78.65 &75.06 & 72.05\\ 
Wong et al., 2020 \cite{wong2020fast}           &    83.34        &   80.25            & 82.03  & 78.81& 74.60 \\
Rebuffi et al., 2021 \cite{rebuffi2021fixing} *   & 83.53    &   81.24  & 82.36     & 80.28& 75.79   \\ 

\hline 
\midrule
\multicolumn{6}{c}{\textbf{WRN-28-10}}\\
\hline 

Baseline &86.67 & 76.17& 0.09& 67.3& 77.33\\
FGSM    \& early stopping & 82.29 &  78.01  &    80.8    &    28.54 &      72.55         \\ 
FGSM \& FLC Pooling (ours)           &  84.93 &81.06 &  83.85 & 72.56&    75.44   \\
Carmon et al., 2019\cite{carmon2019unlabeled} * &\textbf{89.69} & \textbf{87.70}& \textbf{89.12}& \textbf{83.55}& 81.30 \\
Hendrycks et al., 2019 \cite{hendrycks2019using}  &87.11&85.02& 86.47& 80.12&85.02\\
Wang et al., 2020 \cite{Wang2020Improving} * &87.50&85.30&86.74 & 80.65 &\textbf{85.30}\\
Zhang et al., 2021 \cite{zhang2021geometryaware} &89.36 & 87.45& 88.70 & 83.08& 80.11
\\ \bottomrule

\end{tabular}

\label{tab:black_box_cifar10_prn}
\end{center}
\end{table}

\setlength{\tabcolsep}{0.8em}
\begin{table}
\caption[]{Robustness against black box attacks for PRN-18 on CIFAR-100. First against Squares \cite{andriushchenko2020square} with $\epsilon=1/255$ and then against perturbations which were created on the baseline network, meaning transferred perturbations (TP), and the baseline model including early stopping (TPE). As well as the accuracy under common corruptions (CC).}
\scriptsize
\begin{center}
\begin{tabular}{l|lccc|c}
\toprule
Model              & Clean & Squares & TP &TPE & CC\\ \midrule
Baseline & 51.92 &45.74 & 11.13& 23.91& 41.22\\
FGSM    \& early stopping  &52.09  &  45.75  &       23.90 &    10.88 &  41.15             \\ 
FGSM \& FLC Pooling  (ours)           & \textbf{54.66}  &48.85 &  45.59 &    45.31 & \textbf{44.18}  \\
Rice et al., 2020 \cite{rice2020overfitting} &53.83 &\textbf{48.92} & \textbf{45.97}&\textbf{46.11} &43.48\\\bottomrule

\end{tabular}

\label{tab:black_box_cifar100}
\end{center}
\end{table}

 \noindent Table \ref{tab:black_box_cifar10_prn} shows that for PRN-18 models our FLC pooling is consistently able to prevent black box attacks better while maintaining clean accuracy compared to other robust models from RobustBench. For WRN-28-10 models, we see a clear trend that models trained with additional data can achieve higher robustness. This is expected as wider networks can leverage additional data more effectively. One should note that all of these methods require different training schedules which are at least five times slower than ours and additional data which further increases the training time. For example, incorporating the {\it ddpm} dataset into the training increases the amount of training time by a factor of twenty. 
For CIFAR-100 (Table \ref{tab:black_box_cifar100}) our model is on par with \cite{rice2020overfitting}.
\subsection{Corruption Robustness}

To demonstrate that our model generalizes the concept of robustness beyond adversarial examples, we also evaluate it on common corruptions incorporated with CIFAR-C \cite{hendrycks2019robustness}. We compare our model against our baseline as well as other RobustBench \cite{robust_bench} models. Similar to the experiments on black box adversarial attacks we distinguish between models using only CIFAR-10 training data and models using extra-data like {\it ddpm} (marked by *). Table \ref{tab:black_box_cifar10_prn} shows that our FLC pooling, when trained only on CIFAR-10, can outperform other adversarially robust models as well as the baseline in terms of robustness against common corruptions for the PRN-18 architecture. As discussed above, WRN-28-10 models are designed to efficiently leverage additional data. As our model is exclusively trained on the clean CIFAR-10 dataset we can not establish the same robustness as other methods on wide networks. However, we can also see a substantial boost in robustness.
Table \ref{tab:black_box_cifar100} reports the results for CIFAR-100. There we can see that FLC pooling not only boosts clean accuracy but also robust accuracy on common corruptions.

\subsection{Shift-Invariance}
Initially, anti-aliasing in CNNs has also been discussed in the context of shift-invariance \cite{zhang2019making}. Therefore, after evaluating our model against adversarial and common corruptions, we also analyze its behavior under image shifts. We compare our model with the baseline as well as the shift-invariant models from \cite{zhang2019making} and \cite{zou2020delving}. 

FLC pooling can outperform all these specifically designed approaches in terms of consistency under shift, while BlurPooling \cite{zhang2019making} does not outperform the baseline. We assume that BlurPooling is optimized for larger image sizes like ImageNet, 224 by 224 pixels, compared to 32 by 32 pixels for CIFAR-10. The adaptive model from \cite{zou2020delving} is slightly better than the baseline but can not reach the consistency of our model. 

\setlength{\tabcolsep}{1em}
\begin{table} 
\caption[]{Consistency of PRN-18 model prediction under image shifts on CIFAR-10. Each model is trained without AT with the same training schedule (see Appendix \ref{ref:train_schedule} for  details).}
\scriptsize
\begin{center}
\begin{tabular}{l|c|c}
\toprule
Model                & Clean & Consistency under shift \\ \midrule
Baseline             &   94.78    & 86.48                   \\ 
BlurPooling \cite{zhang2019making}         &  \textbf{95.04}     & 86.19                   \\ 
adaptive BlurPooling \cite{zou2020delving}&   94.97    & 91.47                     \\ 
FLC Pooling  (ours)        &   94.66     & \textbf{94.46}                   \\ \bottomrule
\end{tabular}
\label{tab:cifar10_shift}
\end{center}
\end{table}

\section{Discussion \& Conclusions}
The problem of aliasing in CNNs or GANs has recently been widely discussed \cite{durall2020watch,jung2020spectral,karras2021aliasfree}. We contribute to this field by developing a fully aliasing-free down-sampling layer that can be plugged into any down-sampling operation. Previous attempts in this direction are based on blurring before down-sampling. This can help to reduce aliasing but can not eliminate it. With FLC pooling we developed a hyperparameter-free and easy plug-and-play down-sampling which supports CNNs native robustness. Thereby, we can overcome the issue of catastrophic overfitting in single-step AT and provide a path to reliable and fast adversarial robustness. 
We hope that FLC pooling will be used to evolve to fundamentally improved CNNs which do not need to account for aliasing effects anymore.

\bibliographystyle{splncs04}
\bibliography{egbib}
\appendix
\newpage
\color{black}
\section{Appendix}

\subsection{Training Schedules} \label{ref:train_schedule}

\noindent\textbf{CIFAR-10 adversarial training schedule:} For our baseline experiments on CIFAR-10, we used the PRN-18 as well as the WRN-28-10 architecture as they give a good trade-off between complexity and feasibility. For the PRN-18 models, we trained for 300 epochs with a batch size of 512 and a circling learning rate schedule with the maximal learning rate $0.2$ and minimal learning rate $0$. We set the momentum to $0.9$ and weight decay to $5e^{-4}$. The loss is calculated via Cross Entropy Loss and as an optimizer, we use Stochastic Gradient Descent (SGD). For the AT, we used the FGSM attack with an $\epsilon$ of $8/255$ and an $\alpha$ of $10/255$ (in Fast FGSM the attack is computed for step size $\alpha$ once and then projected to $\epsilon$). For the WRN-28-10 we used a similar training schedule as for the PRN-18 models but used only 200 epochs and a smaller maximal learning rate of $0.08$.

\noindent\textbf{CIFAR-10 clean training schedule:} Each model is trained without AT. We used 300 epochs, a batch size of 512 for each training run and a circling learning rate schedule with the maximal learning rate at $0.2$ and minimal at $0$. We set the momentum to $0.9$ and a weight decay to $5e^{-4}$. The loss is calculated via Cross Entropy Loss and as an optimizer, we use Stochastic Gradient Descent (SGD).

\noindent\textbf{CINIC-10 adversarial training schedule:} For our baseline experiments on CINIC-10 we used the PRN-18 architecture. We used 300 epochs, a batch size of 512 for each training run and a circling learning rate schedule with the maximal learning rate at $0.1$ and minimal at $0$. We set the momentum to $0.9$ and weight decay to $5e^{-4}$. The loss is calculated via Cross Entropy Loss and as an optimizer, we use Stochastic Gradient Descent (SGD). For the AT, we used the FGSM attack with an epsilon of $8/255$ and an alpha of $10/255$.

\noindent\textbf{CIFAR-100 adversarial training schedule:} For our baseline experiments on CIFAR-100 we used the PRN-18 architecture as it gives a good trade-off between complexity and feasibility. We used 300 epochs, a batch size of 512 for each training run and a circling learning rate schedule with the maximal learning rate at $0.01$ and minimal at $0$. We set the momentum to $0.9$ and a weight decay to $5e^{-4}$. The loss is calculated via Cross Entropy Loss and as an optimizer, we use Stochastic Gradient Descent (SGD). For the AT, we used the FGSM attack with an epsilon of $8/255$ and an alpha of $10/255$.

\noindent\textbf{ImageNet adversarial training schedule:}
For our experiment on ImageNet we used the ResNet50 architecture.
We trained for 150 epochs with a batch size of 400, and a multistep learning rate schedule with an initial learning rate $0.1$, $\gamma=0.1$, and milestones $[30,60,90,120]$.
We set the momentum to $0.9$ and weight decay to $5e^{-4}$.
The loss is calculated via Cross Entropy Loss and as an optimizer, we use Stochastic Gradient Descent (SGD).
For the AT, we used FGSM attack with an epsilon of $4/255$ and an alpha of $5/255$.

\subsection{ImageNet Training Efficiency}
\label{sec:imagenet_time}

When evaluating practical training times (in minutes) on ImageNet per epoch, we can not see a measurable difference in the costs between a ResNet50 with FLC pooling or strided convolution.

We varied the number of workers for dataloaders with clean training on 4 A-100 GPUs and measured $\approx43$m for 12 workers, $\approx22$m for 48 workers and $\approx18$m for 72 workers for both.
FGSM-based AT with the pipeline by [42] takes 1:07 hours for both FLC pooling and strided convolutions per epoch.
We conclude that training with FLC pooling in terms of practical runtime is scalable (runtime increase in ms-s range) and training times are likely governed by other factors.

 The training time of our model should be comparable to the one from Wong et al.~\cite{wong2020fast} while other reported methods have a significantly longer training time. Yet, the clean accuracy of the proposed model using FLC pooling improves about 8\% over the one reached by \cite{wong2020fast}, with a 1\% improvement in robust accuracy.  For example  \cite{robustness_github} has an increased training time by factor four compared to our model, already on CIFAR10 (see Table \ref{tab:time_report}).
This model achieves overall comparable results to ours.
The model by Salman et al. \cite{salman2020adversarially} is trained with the training schedule from Madry et al. \cite{madry2017towards} and uses a multi-step adversarial attack for training.
Since there is no release of the training script of this model on ImageNet, we can only roughly estimate their training times.
Since they adopt the training schedule from Madry et al., we assume a similar training time increase of a factor of four, which is similar to the multi-step times reported for PGD in Table \ref{tab:time_report}.

\subsection{Aliasing Free Down-Sampling}
\label{sec:free_down}
Previous approaches like \cite{zhang2019making,zou2020delving} have proposed to apply blurring operations before down-sampling, with the purpose of achieving models with improved shift invariance. Therefore, they apply Gaussian blurring directly on the feature maps via convolution.
In the following, we briefly discuss why this setting can not guarantee to prevent aliasing in the feature maps, even if large convolutional kernels would be applied, and why, in contrast, the proposed FLC pooling can guarantee to prevent aliasing.

To prevent aliasing, the feature maps need to be band-limited before down-sampling \cite{Gonzalez}. 
This band limitation is needed to ensure that after down-sampling no replica of the frequency spectrum overlap (see Figure \ref{fig:aliasing_theory}). To guarantee the required band limitation for sub-sampling with a factor of two to $N/2$ where $N$ is the size of the original signal, one has to remove (reduce to zero) all frequency components above $N/2$.

\subsubsection{Spatial Filtering based Approaches} \cite{zhang2019making,zou2020delving} propose to apply approximated Gaussian filter kernels to the feature map. This operation is motivated by the fact that an actual Gaussian in the spatial domain corresponds to a Gaussian in the frequency (e.g.~Fourier) domain. As the standard deviation of the Gaussian in the spatial domain increases, the standard deviation of its frequency representation decreases. Yet, the Gaussian distribution has infinite support, regardless of its standard deviation, i.e.~the function never actually drops to zero. The convolution in the spatial domain corresponds to the point-wise multiplication in the frequency domain. 

Therefore, even after convolving a signal with a perfect Gaussian filter with large standard deviation (and infinite support), all frequency components that were $\neq 0$ before the convolution will be afterwards (although smaller in magnitude). Specifically, the convolution with a Gaussian (even in theoretically ideal settings), can reduce the apparent aliasing but some amount of aliasing will always persist. In practice, these ideal settings are not given: Prior works such as \cite{zhang2019making,zou2020delving} have to employ approximated Gaussian filters with finite support (usually not larger than $7\times 7$). 

\subsubsection{FLC Pooling}
Therefore, FLC pooling operates directly in the frequency domain, where it removes all frequencies that can cause aliases. 

This operation in the Fourier domain is called the \emph{ideal low pass filter} and corresponds to a point-wise multiplication of the Fourier transform of the feature maps with a rectangular pulse $H(\hat{m},\hat{n})$.
\begin{equation}
    H(\hat{m},\hat{n}) =
    \begin{cases}
      1 & \text{for all $\hat{m}$,$\hat{n}$ below M/2 and N/2}\\
      0 & \text{otherwise}
    \end{cases} 
\end{equation}
This trivially guarantees all frequencies above below M/2 and N/2 to be zero.

\subsubsection{Could we apply FLC Pooling as Convolution in the Spatial Domain?}
In the spatial domain, the ideal low pass filter operation from above corresponds to a convolution of the feature maps with the Fourier transform of the rectangular pulse $H(\hat{m},\hat{n})$ (by the Convolution Theorem, e.g.\cite{Gonzalez}). 
The Fourier transform of the rectangle function is
\begin{equation}
sinc(m,n) = 
\begin{cases}
\frac{sin(\sqrt{m^2+n^2})}{\sqrt{m^2+n^2}} & m, n \neq 0 \\
1 & m,n =0
\end{cases}
\end{equation}
 However, while the ideal low pass filter in the Fourier domain has finite support, specifically all frequencies above $N/2$ are zero, $sinc(m,n)$ in the spatial domain has infinite support. 
 Hence, we need an infinitely large convolution kernel to apply perfect low pass filtering in the spatial domain. This is obviously not possible in practice. In CNNs the standard kernel size is 3x3 and one hardly applies kernels larger than 7x7 in CNNs.

\subsection{Model Confidences}
In Table \ref{tab:confidence_cifar10}, we evaluate the confidence of model predictions. We compare each model's confidence on correctly classified clean examples to its respective confidence on wrongly classified adversarial examples. Ideally, the confidence on the adversarial examples should be lower. The results for the different methods show that FLC yields comparably high confidence on correctly classified clean examples with a 20\% gap in confidence to wrongly classified adversarial examples. In contrast, the baseline model is highly confident in both cases. Other, even state-of-the-art robustness models have on average lower confidences but are even less confident in their correct predictions on clean examples than on erroneously classified adversarial examples (e.g.~MART~\cite{Wang2020Improving} and PGD~\cite{pgd}). Only the model from \cite{wu2020adversarial} has a trade-off preferable over the one from the proposed, FLC  model.

\begin{table}[t]
\small
\caption[]{Evaluation for clean and robust accuracy, higher is better, on AutoAttack \cite{auto_attack} with our trained models. The models reported by the original authors may have different numbers due to different hyperparameter selection. We report each models confidence on their correct predictions on the clean data (Clean Confidence) and the models confidence on its false predictions due to adversarial perturbations (Perturbation Confidence). The top row reports the baseline without adversarial training.}
\small
\begin{center}
\begin{adjustbox}{width=1\textwidth}
\begin{tabular}{l|cc|cc}
\toprule
Method      & Clean Acc & AA Acc   & \makecell{Clean \\Confidence}& \makecell{Perturbation \\Confidence}   \\ \midrule
Baseline & 95.08 & 0.00 & 100.00 & 97.89 \\ \midrule
FGSM    \& early stopping \cite{wong2020fast}  & 82.88       & 11.82 & 90.50 &84.26   \\ 
FGSM    \& FLC Pooling (Ours) &  \textbf{84.81} & 38.41 & \textbf{98.84}&70.98 \\ 
PGD \cite{pgd} &   83.11 & 40.35 & 56.58& 75.00  \\ 
Robustness lib \cite{robustness_github} &  76.37 & 32.10 &95.22 & 78.91\\
AWP \cite{wu2020adversarial}&  82.61 & \textbf{49.43} & 88.83&\textbf{37.98}\\ 
MART \cite{Wang2020Improving} & 55.49 & 8.63 &24.44 &50.17\\
TRADES \cite{trades} &     81.49 & 46.91 &53.94 & 50.46\\ \bottomrule
\end{tabular}
\end{adjustbox}
\label{tab:confidence_cifar10}
\end{center}
\end{table}

\subsection{AutoAttack Attack Structure}
\begin{figure}
 \tiny
\begin{center}
\begin{adjustbox}{width=0.95\textwidth}
\begin{tabular}{cc}
\begin{minipage}{.19\textwidth}
      \includegraphics[width=\linewidth]{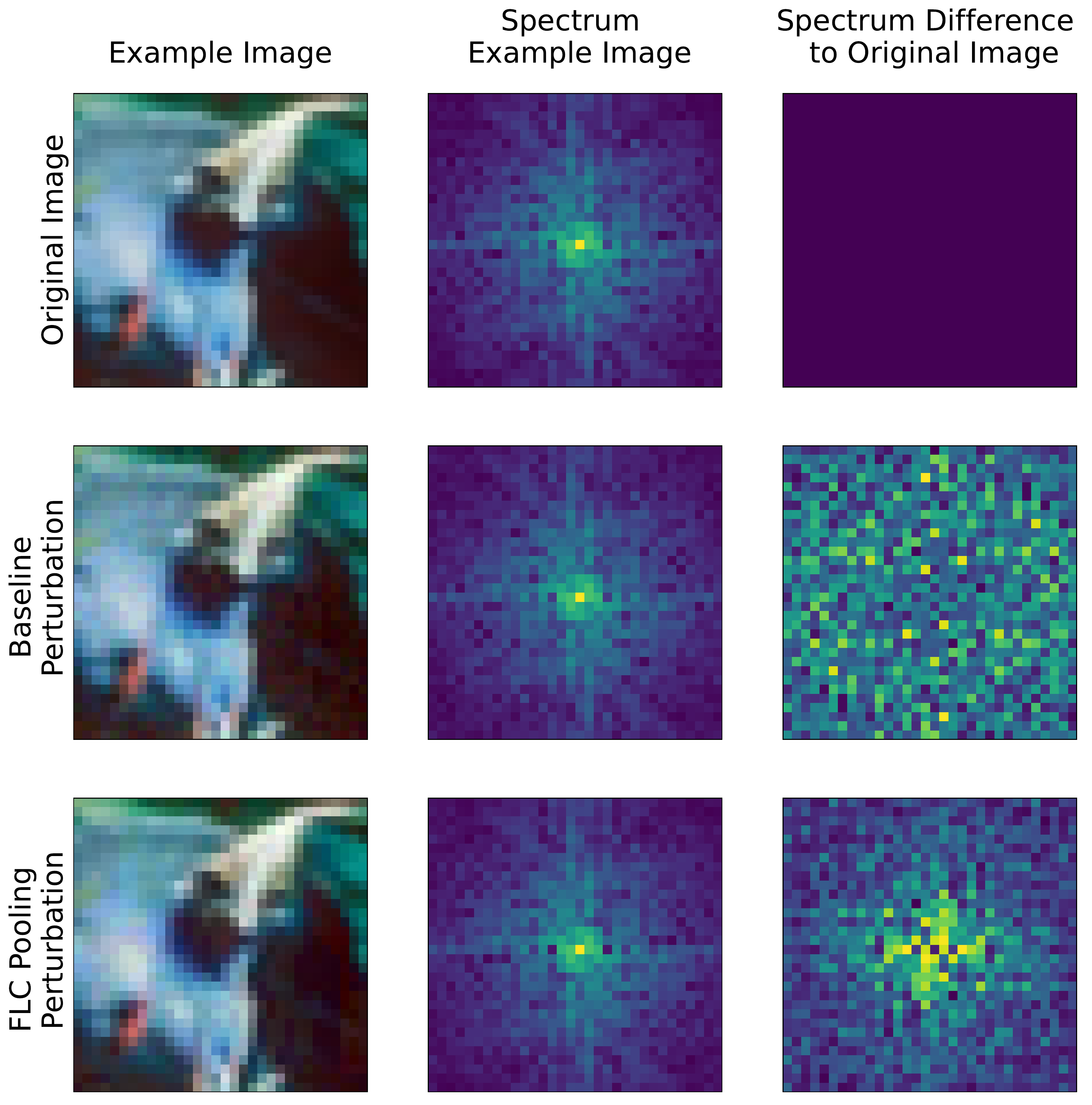}
    \end{minipage} &
\begin{minipage}{.19\textwidth}
      \includegraphics[width=\linewidth]{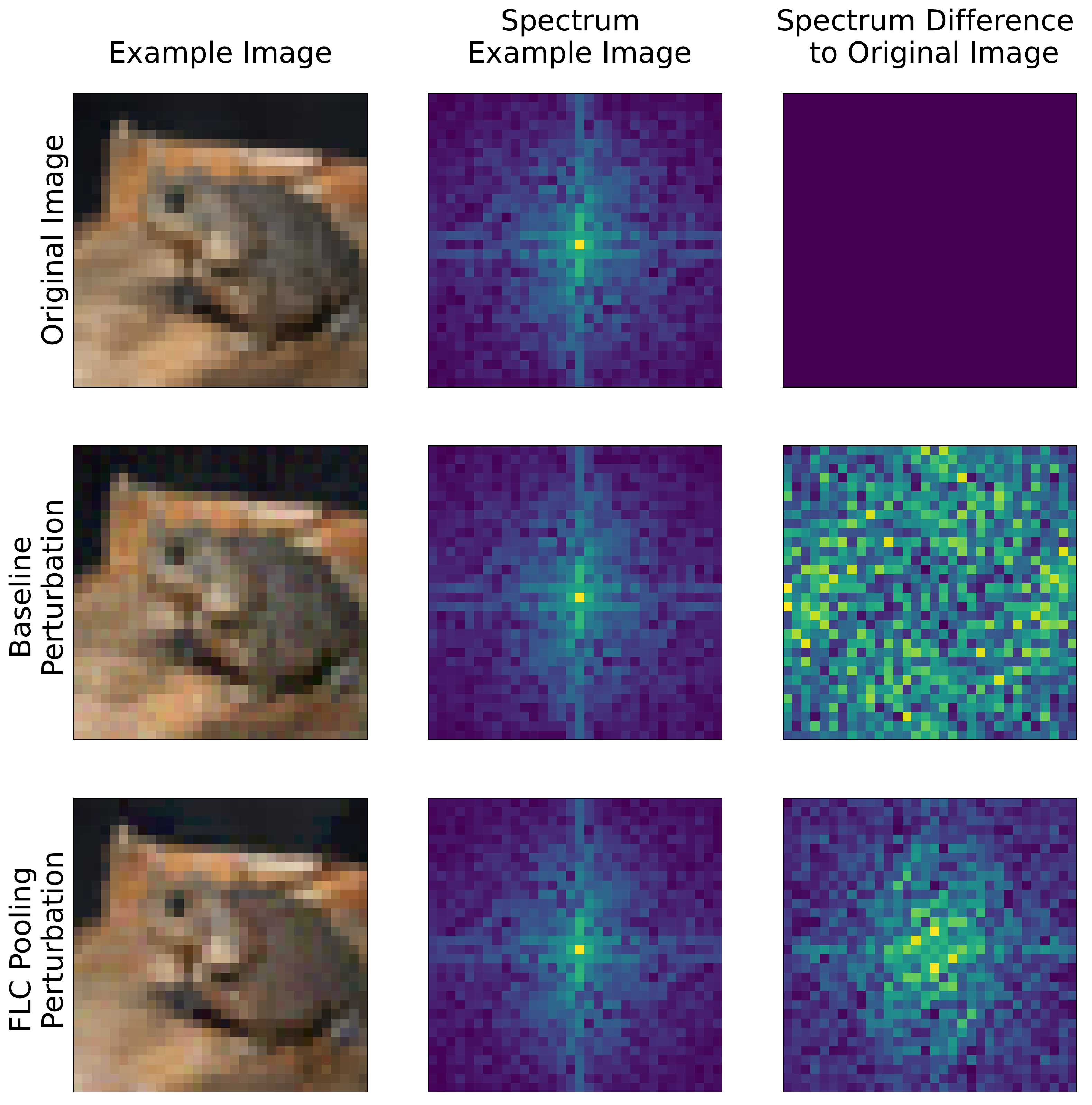}
     \end{minipage}   \\
  \begin{minipage}{.19\textwidth}
      \includegraphics[width=\linewidth]{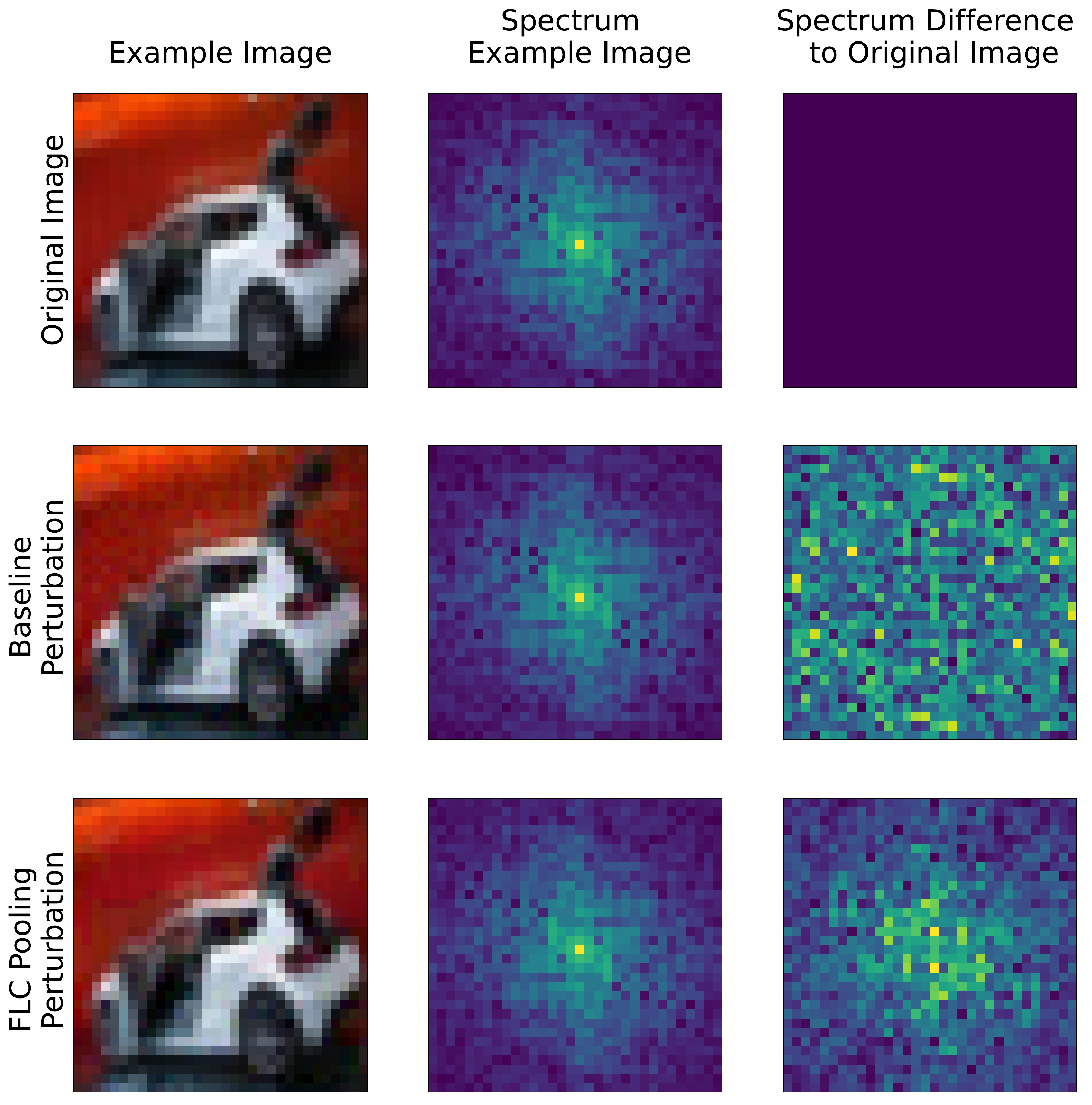}
    \end{minipage} &
\begin{minipage}{.19\textwidth}
      \includegraphics[width=\linewidth]{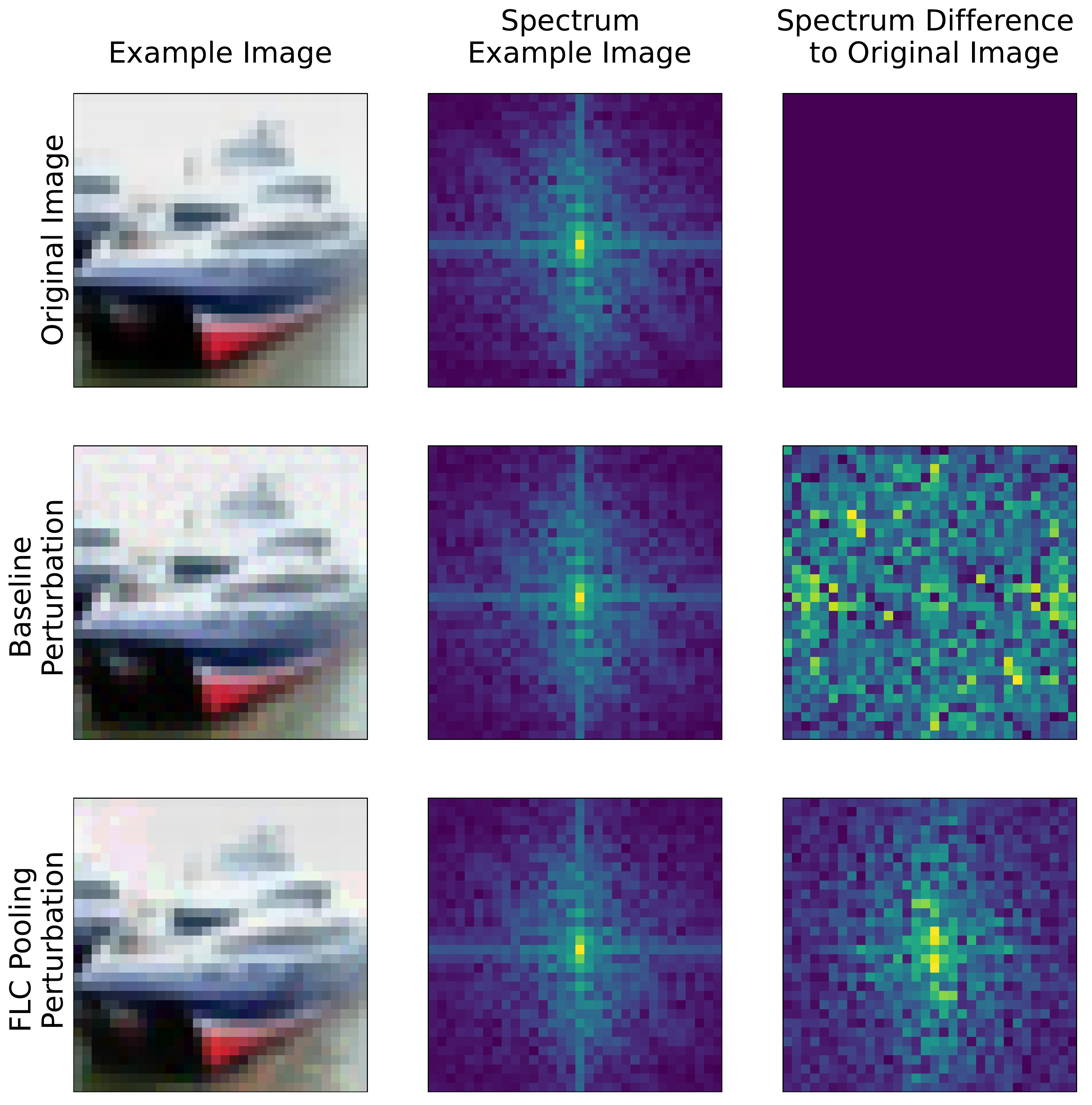}
     \end{minipage}   \\   
      \begin{minipage}{.19\textwidth}
      \includegraphics[width=\linewidth]{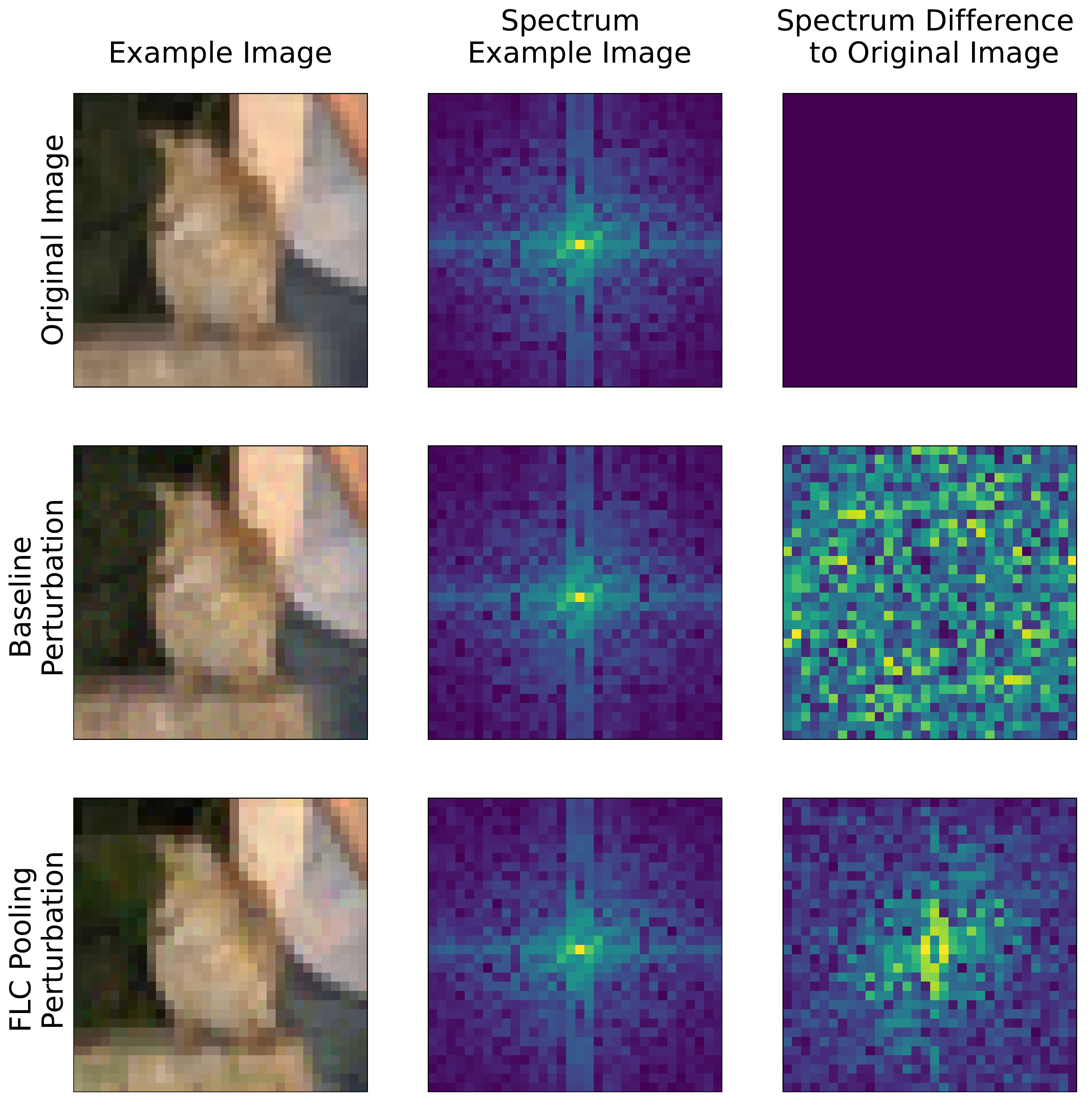}
    \end{minipage} &
\begin{minipage}{.19\textwidth}
      \includegraphics[width=\linewidth]{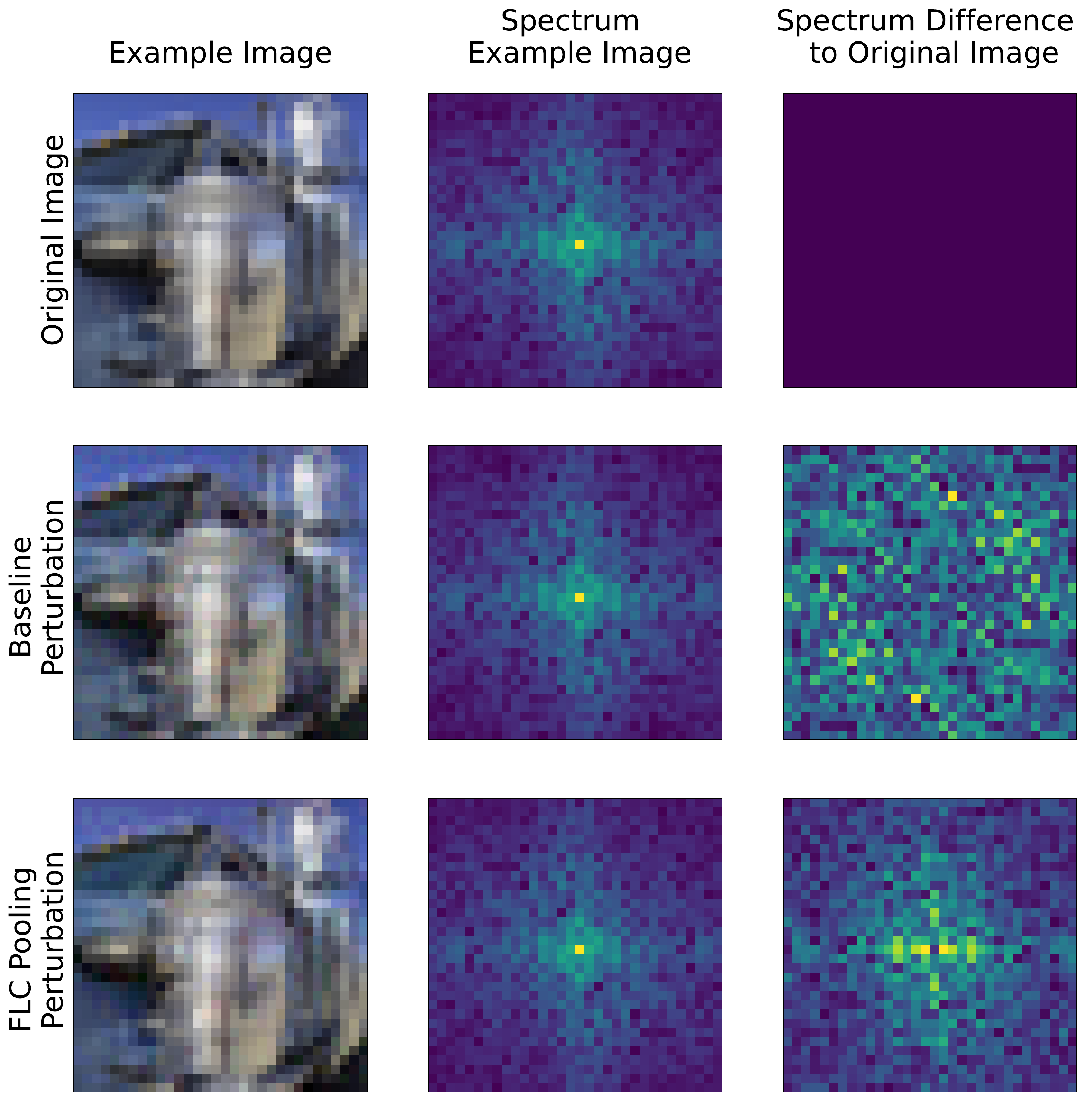}
     \end{minipage}   \\  
     
 \end{tabular}
 \end{adjustbox}
 \caption[]{Spectrum and spectral differences of adversarial perturbations created by AutoAttack with $\epsilon=\frac{8}{255}$ on the baseline model as well as our FLC Pooling. The classes from top left down to the bottom right are: Bird, Frog, Automobile, Ship, Cat and Truck.}
 \label{fig:attack_images_appendix}
 \end{center}
 \end{figure}

In the main paper we showed one example of an image optimized by 
 AutoAttack~\cite{auto_attack} to fool our model and the baseline  in Figure \ref{fig:attack_images}. In Figure \ref{fig:attack_images_appendix}, we give more examples for better visualisation and comparison.

\subsection{Ablation Study: Additional Frequency Components}

\begin{figure*}[ht]
 \scriptsize
\begin{center}

\begin{tabular}{@{}c@{}c@{}}
\begin{minipage}{.45\textwidth}
      \includegraphics[width=\linewidth]{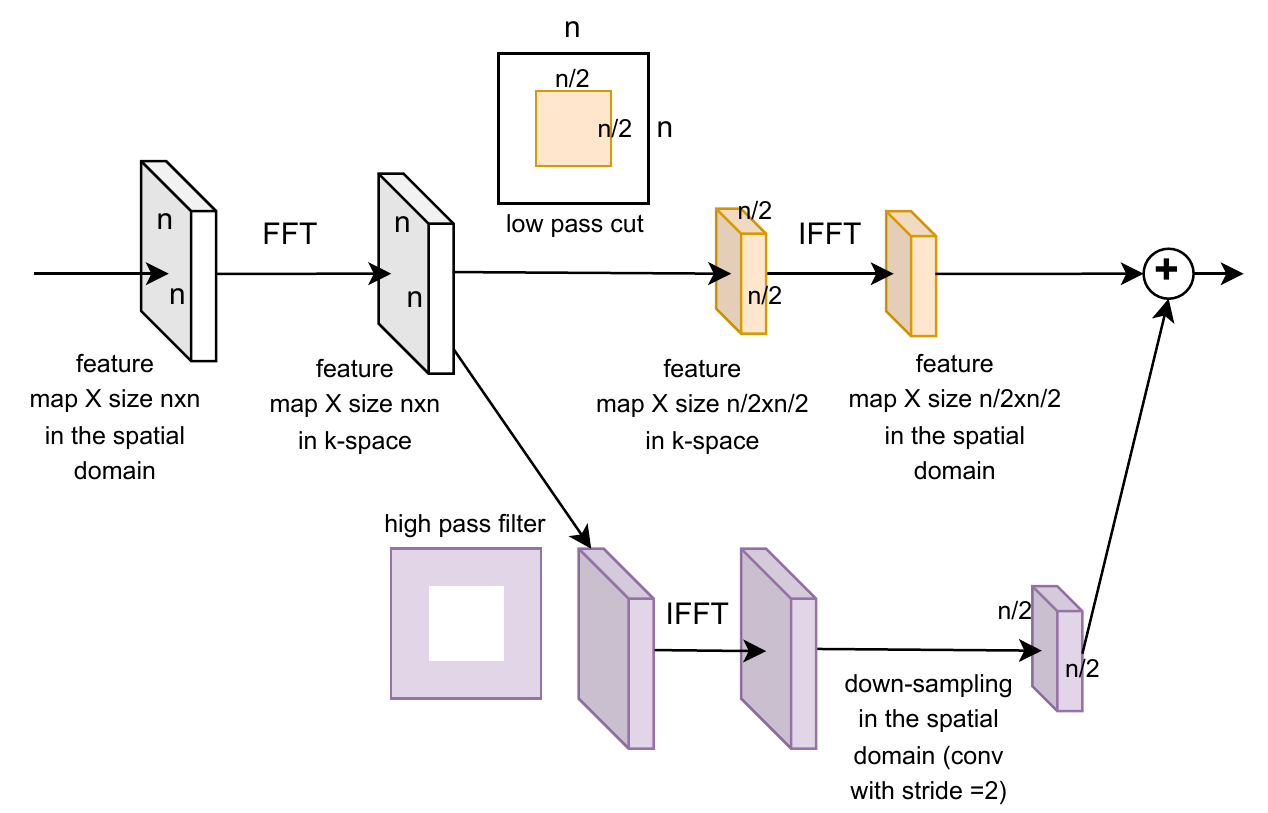}
     \end{minipage}  

&
  \begin{minipage}{.45\textwidth}
      \includegraphics[width=\linewidth]{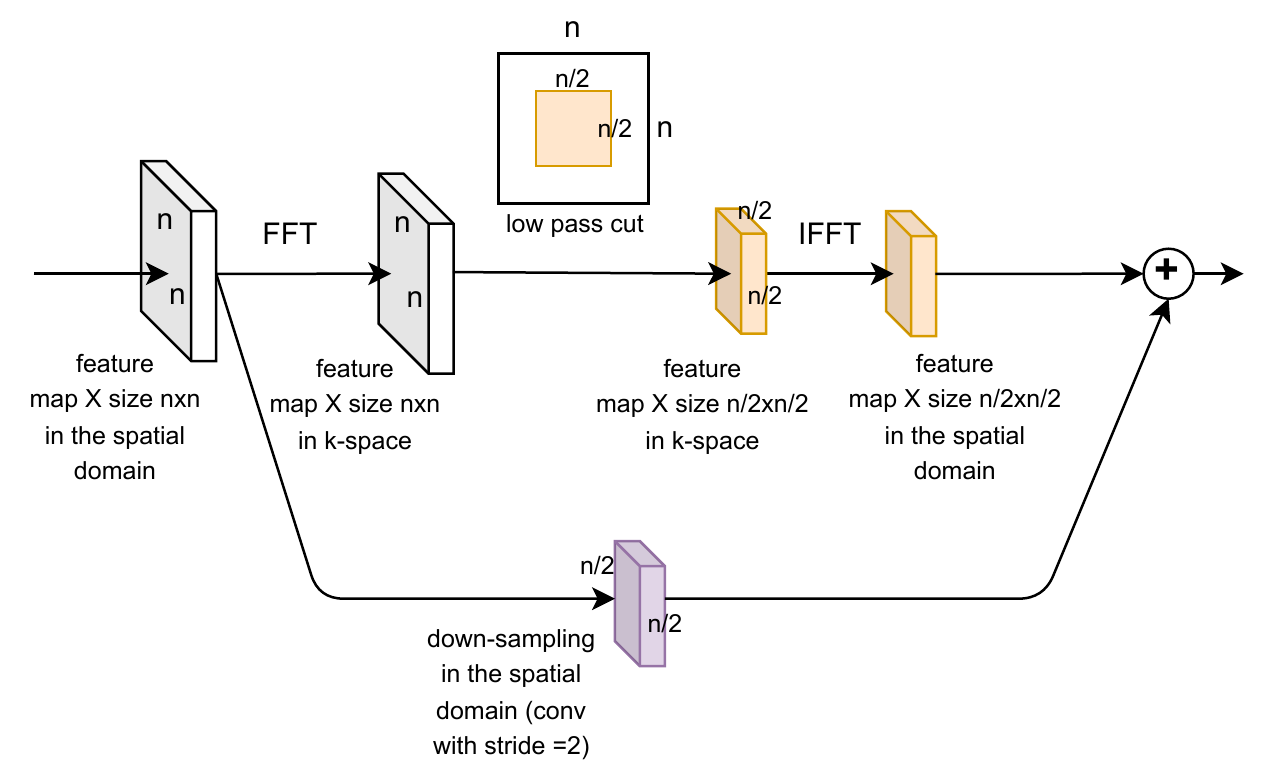}
     \end{minipage}  
 
 \end{tabular}
 \\
 \caption[]{LC pooling plus, which either includes the original down-sampled signal like it is done traditionally (right) or with the high frequency components filtered by a high pass filter in the Fourier domain and down-sampled in the spatial domain by an identity convolution of stride two (left).}
 \label{fig:full_path}
 \end{center}
 \end{figure*}

In addition to the low frequency components we tested different settings in which we establish a second path through which we aim to add high frequency or the original information. We either add up the feature maps or contacted them. The procedure of how to include a second path is represented in Figure \ref{fig:full_path}. One approach is to execute the standard down-sampling and add it to the FLC pooled feature map. The other is to perform a high pass filter on the feature map and down-sample these feature maps. Afterwards, the FLC pooled feature maps as well as the high pass filtered and down-sampled ones are added. With this ablation, we aim to see if we do lose too much through the aggressive FLC pooling and if we would need additional high frequency information which is discarded through the FLC pooling.
Table \ref{tab:cifar10lowplus} show that we can gain minor points for the clean but not for the robust accuracy. Hence we did not see any improvement in the robustness and an increase in training time per epoch as well as a minor increase in model size, we will stick to the simple FLC pooling.

\begin{table}
\scriptsize
\caption[]{Accuracies for CIFAR-10 Baseline LowCutPooling plus the original or high freqeuncy part of the featuremaps down-sampled in the spatial domain for FGSM Training. We can see that the additional data does not improve the robust accuracy and gives only minor improvement for the clean accuracy. Due to the additional computations necessary for the high frequency /original part we decided to fully discard them and stick to the pure low frequency cutting.}
\begin{center}
\begin{tabular}{l|cc|c|c}
\toprule
Method                             & \makecell{Clean }  & \makecell{PGD $L_{\inf}$ \\ $\epsilon=\frac{8}{255}$} & \makecell{Seconds per \\ epoch (avg)} & \makecell{Model size \\(MB)}\\ \midrule
FLC pooling                         &  84.81    &  38.41   &  34.6 $\pm$ 0.1     & 42.648 \\ 
FLC pooling + HighPass pooling          & 85.38    & 38.02   &  45.2 $\pm$ 0.4   & 42.652 \\ 
FLC pooling + Original pooling      &  85.37    & 38.30   &    35.4 $\pm$ 0.1 & 42.652  \\ \bottomrule
\end{tabular}
\label{tab:cifar10lowplus}
\end{center}
\end{table}

\vfill
\pagebreak

\end{document}